\documentclass{article}
\usepackage{arxiv}

\usepackage[utf8]{inputenc} 
\usepackage[T1]{fontenc}    
\usepackage{hyperref}       
\usepackage{url}            
\usepackage{booktabs}       
\usepackage{amsfonts}       
\usepackage{nicefrac}       
\usepackage{microtype}      
\usepackage{lipsum}		
\usepackage{graphicx}
\usepackage{natbib}
\usepackage{doi}
\usepackage{amsmath}
\usepackage[capitalise]{cleveref}

\title{The Artificial Intelligence behind the winning entry to the 2019 AI Robotic Racing Competition}

\date{} 					

\author{
\href{https://orcid.org/0000-0002-6795-8454}{\includegraphics[scale=0.06]{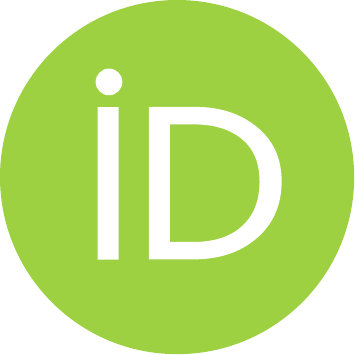}\hspace{1mm}Christophe De~Wagter}\thanks{http://mavlab.tudelft.nl.} \\
Micro Air Vehicle Lab\\
Delft University of Technology\\
2629HS Delft \\
the Netherlands \\
\texttt{c.dewagter@tudelft.nl} \\
\And
\href{https://orcid.org/0000-0002-9478-7195}{\includegraphics[scale=0.06]{orcid}\hspace{1mm}Federico Paredes-Vallés} \\
Micro Air Vehicle Lab\\
Delft University of Technology\\
2629HS Delft \\
the Netherlands \\
\texttt{F.ParedesValles@tudelft.nl} \\
\AND
Nilay Sheth \\
Micro Air Vehicle Lab\\
Delft University of Technology\\
2629HS Delft \\
the Netherlands \\
\texttt{nilay\_994@hotmail.com} \\
\And
\href{https://orcid.org/0000-0001-8265-1496}{\includegraphics[scale=0.06]{orcid}\hspace{1mm}Guido C. H. E. de~Croon} \\
Micro Air Vehicle Lab\\
Delft University of Technology\\
2629HS Delft \\
the Netherlands \\
\texttt{g.c.h.e.decroon@tudelft.nl}
}

\renewcommand{\shorttitle}{The AI behind the winning entry to the 2019 AIRR}

\hypersetup{
pdftitle={The Artificial Intelligence behind the winning entry to the 2019 AI Robotic Racing Competition},
pdfsubject={Autonomous drone racing},
pdfauthor={C. De~Wagter, F. Paredes-Vallés, N. Sheth, G. C. H. E. de~Croon},
pdfkeywords={Autonomous drone racing, AI},
}

\begin{document}

\maketitle

\begin{abstract}
Robotics is the next frontier in the progress of Artificial Intelligence (AI), as the real world in which robots operate represents an enormous, complex, continuous state space with inherent real-time requirements. One extreme challenge in robotics is currently formed by autonomous drone racing. Human drone racers can fly through complex tracks at speeds of up to 190 km/h. Achieving similar speeds with autonomous drones signifies tackling fundamental problems in AI under extreme restrictions in terms of resources. In this article, we present the winning solution of the first AI Robotic Racing (AIRR) Circuit, a competition consisting of four races in which all participating teams used the same drone, to which they had limited access. The core of our approach is inspired by how human pilots combine noisy observations of the race gates with their mental model of the drone's dynamics to achieve fast control. Our approach has a large focus on gate detection with an efficient deep neural segmentation network and active vision. Further, we make contributions to robust state estimation and risk-based control. This allowed us to reach speeds of ~9.2m/s in the last race, unrivaled by previous autonomous drone race competitions. Although our solution was the fastest and most robust, it still lost against one of the best human pilots, Gab707. The presented approach indicates a promising direction to close the gap with human drone pilots, forming an important step in bringing AI to the real world.
\end{abstract}


\section{Introduction}


Artificial Intelligence has seen tremendous progress over the last decade, especially due to the advent of deep neural networks \cite{krizhevsky2012imagenet,schmidhuber2015deep}. The major milestones in the history of AI have always been associated with competitions against human experts. These competitions clearly show the increasing complexity in the tasks in which AI can extend beyond human performance. In 1997, IBM’s Deep Blue showed the power of search methods combined with expert systems \cite{campbell2002deep} by beating the world champion in the game of chess, Garry Kasparov. Chess is a fully observable, turn-based game, with ~10123 possible game states. After chess, the AI community started to aim for the game of Go, which has a much larger branching factor that also results in a much higher number of ~10360 possible game states, rendering most search methods ineffective. In 2017, the Master version of Google Deepmind’s AlphaGo beat Ke Jie, the top-ranked Go player at the time. AlphaGo used an elegant combination of Monte Carlo tree search and deep neural networks for evaluating board positions \cite{silver2016mastering}. In 2019, Google Deepmind’s AlphaStar reached a GrandMaster status in the real-time strategy game StarCraft II \cite{vinyals2019grandmaster}. This game represents yet a higher complexity, as it involves real-time instead of turn-based play, partial observability, and a large and varied action space. 

\begin{figure}[hbt!]
    \centering
    \includegraphics[width=\textwidth]{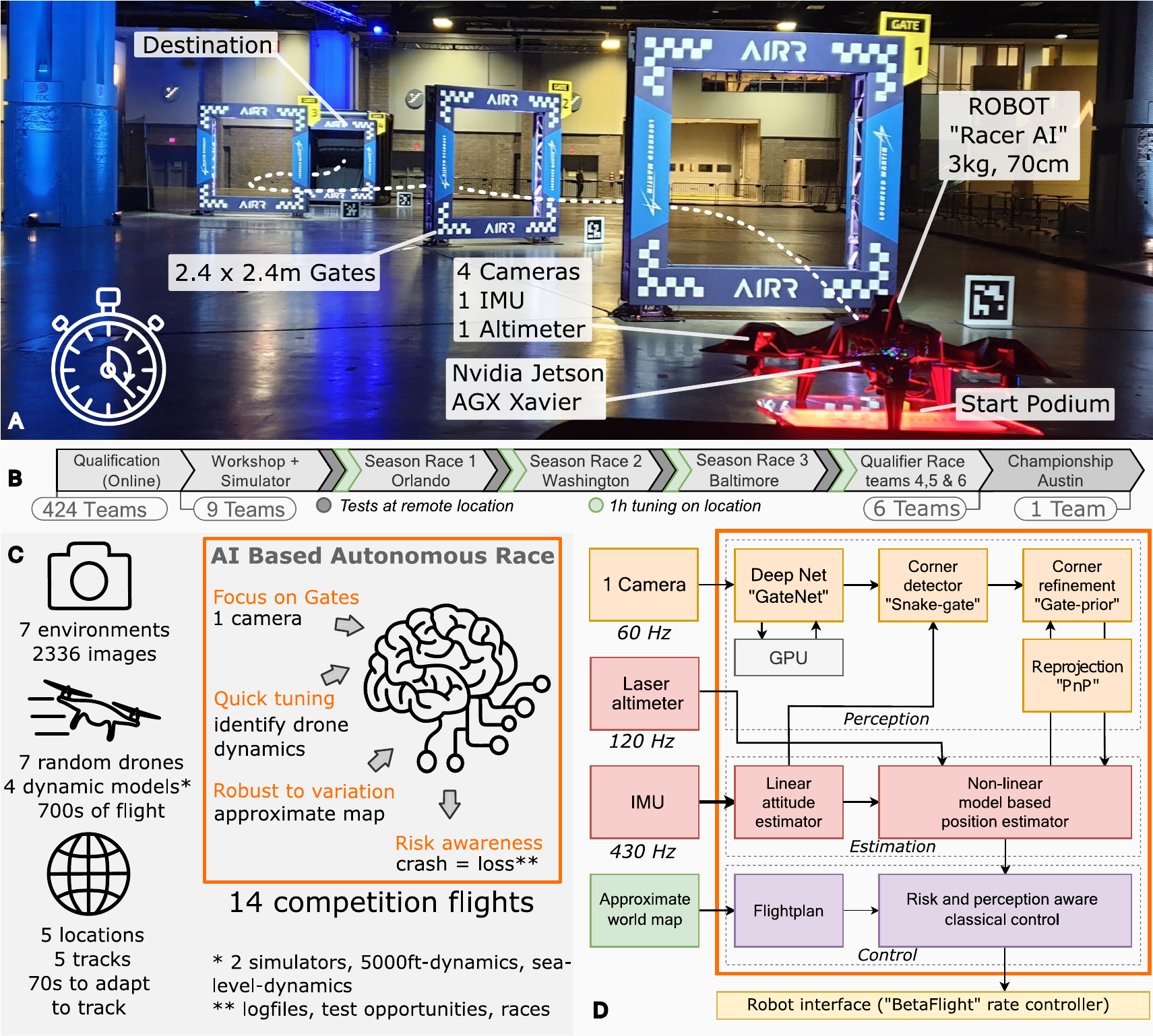}
    \caption{\textbf{Human inspired approach to drone racing: single-camera position estimation, dynamics prediction, a rough model of the track.} \textbf{(A)} Race elements. \textbf{(B)} Timeline of the season with an online qualification phase, 3 season races, and a championship race. \textbf{(C)} Training data overview and approach to focus on gate detection, quick tuning, robustness to variation, and risk awareness. \textbf{(D)} System schematic of the approach.}
    \label{fig1}
\end{figure}

Robotics will form a new frontier in AI research since the associated problems are even more complex \cite{yang2018grand}. Typical robotics problems are high-dimensional, continuous, and only partially observable. Moreover, and most importantly, robots have to operate in the real world, of which many relevant aspects remain hard to model or simulate. Sample-intensive learning methods may apply to simplified robot models in simulation, allowing for faster than real-time learning, but transferring them to an actual robotic system typically leads to a reality gap \cite{jakobi1995noise,koos2012transferability,scheper2016abstraction} that substantially reduces performance. One part of the reality gap is the difference in sensory input like visual appearance and sensor noise. The other part of this reality gap is the specifics of a robot itself, concerning both its ``body'' (energy source, structure, sensors, actuators) and ``brain'' (processing power, memory). For example, there may be unmodeled aerodynamic effects or different timings in the perception-action cycle of the actual robotic hardware.

One extreme challenge at the moment for AI in robotics is formed by autonomous drone racing. Similar to human drone races, the goal for the drones is to finish a pre-determined racing track in as short a time as possible (see \cref{fig1}A). The drones have to race by using only their onboard resources, such as onboard sensors and processing, which are heavily restricted in terms of size, weight, and power (SWaP) \cite{Floreano2015}. To be successful, the drones will have to fly through complex tracks at very high speeds (human racers reach speeds of up to 190km/h). This means that they need a fast perception-action cycle on light-weight hardware, which additionally should be robust, as the margin for error is small. 

The research on autonomous drone racing finds its roots in seminal work on agile and aggressive flight \cite{lupashin2010simple,mellinger2012trajectory,mellinger2011minimum}. The focus of many of these early studies was mostly on high-performance control, outsourcing sensing, and state estimation to external motion tracking systems and associated central computers. Later studies focused on getting also the sensing and state estimation onboard, allowing the drones to perform quick maneuvers through gaps \cite{falanga2017aggressive,loianno2016estimation}. A real drone race additionally requires the drone to detect racing gates in more complex spaces, with multiple gates and potential distractors in view, while not only passing one gate, but flying a full trajectory in sequence, dealing with unforeseen deviations on the way. The research on drone racing received a boost by the first-ever autonomous drone race competition, organized in conjunction with the IROS robotics conference in 2016, in Daejeon, South Korea \cite{moon2017iros}. This competition let the participants free in their choice of platform and only required that all sensing and processing took place onboard. The first competition showed the difficulty of the problem, with the winner reaching 10 gates at an average speed of 0.6 m/s. This is in stark contrast to the impressive racing performance reached a year later by Morrell et al. \cite{morrell2018differential}, whose drone only lost by a few seconds from an expert human pilot on their track. In-competition flight speeds remained inferior to out-of-competition flight speeds also over the ensuing years, with IROS drone race speeds of the winner reaching 2.5 m/s in 2018 \cite{kaufmann2019beauty}. The reason for this mostly lies in the real-world aspects of the competitions. They take place in environments previously unknown by the teams, with no opportunity for benign, solution-specific changes, and little time for adapting the developed solution to the environment in situ. Moreover, competitions often pose a more challenging environment, with gates located slightly differently than on the pre-communicated maps or even moving during the race, unforeseen lighting effects optimized for spectators rather than for drones, and large crowds of moving people around the flight arena.

In this article, we present our winning approach to the 2019 Artificial Intelligence Robotic Racing (AIRR) competition (see \cref{fig1}), which was organized by Lockheed Martin and the Drone Racing League (DRL) in 2019 and had a grand prize for the best AI of 1M\$. The AIRR competition strives to support the development of AI for racing drones that will be able to surpass human drone racing pilots. It is completely different from the previous autonomous drone racing competitions in many aspects. For example, the competition did not take place in a single day at a conference but had two phases: a qualifier phase and a competition phase. In the qualifier phase, 424 teams registered worldwide and had to qualify by performing a computer vision task, racing in simulation, and describing their proposed approach and team composition. The competition phase was also unique as it was organized as a complete season with three seasonal races and a championship race. The races themselves were organized by DRL as e-sports events, also aiming for the amusement of the audience, and adding specific requirements on the teams and robots. Moreover, the organization provided all teams with the same type of racing drone, developed by the organizers. These drones were equipped with four high-resolution, wide field-of-view cameras, and an NVIDIA Xavier board to run the embedded AI (see \cref{fig1}A). Hence, the robotic hardware was the same for all teams, making the competition only about the difference between the AI software. Moreover, the teams had very little direct access to the racing drone hardware, making it very hard to get acquainted with the hardware, to perform calibration, and to identify potential reality gaps. The amount of flight testing was low and happened in different conditions than the races in terms of light, room size, and even air density. Finally, during the races, the AI code was uploaded to refurbished drones that had never flown this particular code before without the possibility for changes based on previous runs. Consequently, the AI developed for the competition had to be very data efficient and robust. The solutions had to mainly be developed on the provided simulator, which figured a substantial reality gap in terms of drone dynamics and sampling characteristics of especially the camera. The simulator did have a hardware-in-the-loop setup for the processing, i.e., it communicated with the same NVIDIA Xavier board and allowed to accurately test the computational effort of the developed algorithms.

\section{Strategy: a human-inspired, gate-based approach to autonomous drone racing}

In our approach to developing an AI for the AIRR competition, we used the characteristics and restrictions of the competition as a point of departure (\cref{fig1}A,B,C). First and foremost, we desired to fly as fast as possible, ideally as close as possible to the physical limits of the drone. This implied that we did not use perception methods that would restrict the drone’s maximum speed. Importantly, it meant the exclusion of state-of-the-art methods for feature-based Visual Inertial Odometry (VIO) (cf., e.g., \cite{delmerico2018benchmark}), since the blurry images that occur at higher speeds lead to more difficulties in finding and tracking of features. The reliance on this type of VIO was one of the main reasons that the runner-up team limited their velocity to 8 m/s \cite{foehn2020alphapilot}. Moreover, current accurate VIO methods have the disadvantage that they are computationally intensive. For similar reasons, we did not employ feature-based Simultaneous Localization And Mapping (SLAM) methods, e.g., \cite{mur2015orb} as used by the winning team in the IROS 2017 competition \cite{moon2019challenges}, where additionally SLAM methods have difficulties handling changes in the map, like the foreseen gate displacements. 

Instead, we drew inspiration from human pilots who focus greatly on the gates, while combining their observations with their knowledge of the drone’s responses to control inputs and an approximate map of the track (see \cref{fig1}C). Hence, we developed an accurate, robust, and computationally efficient monocular gate detection method. We aimed to process images at frame-rate at the fastest speeds the drone could achieve. Whereas previous competitions contained gates of a uniform, unique color \cite{moon2017iros,moon2019challenges}, the AIRR competition featured more complex gates, precluding hand-designed detection methods as in \cite{li2020autonomous,li2020visual}. Relative localization can also not be done with standard rectangle-based detectors such as SSD \cite{jung2018perception} since the bounding boxes generated by such methods by themselves do not allow for an accurate determination of the drone’s relative pose. Furthermore, we did not choose for a deep neural network that immediately maps images to relative pose, as in \cite{cocoma2019cnn,kaufmann2019beauty}. Such networks experience difficulties when multiple gates are in sight and are more difficult to analyze and fix.

We used a gate segmentation deep neural network called ``GateNet'' to create a fast vision pipeline that is minimally sensitive to increasing levels of motion blur, rolling shutter deformation, the possible absence of texture in large parts of the scene, or even the presence of moving unknown entities. Subsequently finding gate corners was done with very efficient active perception \cite{li2020autonomous}. Especially on flying robots where every gram matters, active vision is highly relevant \cite{Bajcsy2018} and we show it can be part of successful engineering designs. Pose estimates finally can then be computed using perspective-n-point (PnP) (see \cref{fig2}D) combined with the racecourse map.

Further inspired by human FPV pilots’ ability to predict drone motion, we enhanced the drone’s state estimates with model-based predictions from a dynamic model fitted on flight data. Merging the visual measurements with the predictions is then done with a random sample consensus (RANSAC) based Moving Horizon Estimator (MHE) from previous work \cite{li2020visual} but extended to better estimate the drone’s yaw angle during the race.

Concerning control, we designed a strategy that would allow for high speeds but would allow flying very early on and would have short intuitive tuning cycles given the little available flight time. As a result, promising methods such as deep reinforcement learning \cite{loquercio2019deep} or imitation learning \cite{rojas2020deeppilot} were ruled out. The short timeline and little flight time would not allow for a thorough investigation of the reality gap between the drone and the simulator with methods like abstraction \cite{scheper2016abstraction,kaufmann2020deep}. Even online adaptation \cite{Johnson2005} would yield limited benefits in a race where every drone only flew once, measurements could be very noisy and the time to the first gate is 2 seconds. Finally, the long downtimes, loss of log files (stored in RAM) and failed competition runs in case of a crash, made risk management a crucial part of the control development. While perception-aware trajectory generation \cite{Murali2019} can optimize speed and perception, it does not take into account that collision risks depend on relative position to the gate.

The control is therefore designed from a gate-centered perspective in which risks and constraints but also position uncertainty vary depending on the distance to the gate. The controller makes use of classical control theory but was gradually augmented to fly increasingly close to the platform limits. This allowed us to start flying early in the process and gather crucial log-files to steadily investigate drone limitations while minimizing risks.

On top of the initial scheme, we implemented an open-loop full-throttle take-off called ``Boost'' to overcome sensor boot-time delays. We adopted a pitch-for-altitude controller to maintain altitude when thrust saturated and an offline optimized gate-approach-line strategy. Finally, we developed a human-inspired risk-aware strategy that speeds the robot up substantially when far from obstacles or when it is aligned with the next gate but slows down when uncertain or misaligned. By reaching various hardware limitations we expect it to be on a par with optimal control methods that employ multiple simplifications (like minimizing snap \cite{mellinger2011minimum} instead of time) or have short time horizons to allow real-time computation onboard of a drone \cite{Bangura2014}.
Whereas both in computer vision and robotics a lot of research effort is invested in increasing the accuracy of methods, we put computational efficiency at the core of our approach. The reason for this is that control performance is not only determined by accuracy but also by the control delay, two factors which are most often on a trade-off with each other. Moreover, not saturating processing power allowed us to have additional threads logging all data (images, states, measurements, etc.). This logged data was extremely important to estimate the drone’s model and fusion parameters and for retraining and improving the perception pipeline.

In the next section we give an in-depth explanation of the 
implementation used in the competition and in our approach.

\section{Implementation}

\subsection{Drone specifications}

All participating teams in the 2019 AIRR competition operated the same race drone type called ``Racer AI'' (see \cref{fig1}A). This plus-configuration quadcopter was approximately 70 cm in diameter, weighed around 3 kg, and had a thrust-to-weight ratio of about 1.4. It was equipped with two sets of ±30-degree forward-facing stereo camera pairs. The cameras were the global-shutter, color Sony IMX 264 sensor, which provided 1200x720 resolution images at a rate of 60 Hz. Besides cameras, the Racer AI had a Bosch BMI088 IMU, characterized by a measurement range of ±24 g and ±34.5 rad/s (with a resolution of 7e-4 g and 1e-3 rad/s) provided at an update rate of 430 Hz; and a downward-facing laser rangefinder Garmin LIDAR-Lite v3 with a range between 1 and 40 m (resolution of 0.01 m) and update rate of 120 Hz. As the embedded computer, the Racer AI was equipped with an NVIDIA Jetson Xavier, containing a GPU with 512 CUDA cores and an 8-core ARM CPU. It ran Linux with the PREEMPT RT kernel patch. Lastly, the Racer AI had a BetaFlight low-level autopilot controlling the angular velocities of the drone and accepting commands at a rate of 50 Hz. 

\subsection{Perception}

The perception modules were executed sequentially in a dedicated thread, while a separate thread did the logging of images. We only used one out of the four cameras, as this setup matches the challenge that human pilots have to face. Although monocular vision is more challenging in terms of depth perception, it entails less computational load and calibration requirements. Since none of the cameras faced forward, we had the drone fly in the direction of the optical axis of the selected camera. The original 1200x720 images provided by the camera were first centrally cropped to 720x720 to remove parts of the own robot that were in sight (See \cref{figS2}A), and then down-sampled to 360x360.

\subsubsection{GateNet: Gate detection by semantic segmentation}

In the first stage of our perception pipeline, GateNet was used to transform each (not undistorted) input image $\mathbf{I}$ into a binary mask $\mathbf{M}$ that segments all visible gates, regardless of their distance to the camera. GateNet is a fully convolutional deep neural network architecture that consists of a 4-level U-Net \cite{Ronneberger2015a} with [64, 128, 256, 256] convolutional filters of size 3x3 and (elementwise-sum) skip connections. All layers use ReLU activation functions except the final prediction layer, which uses a sigmoid to keep $\mathbf{M}$ in the range [0, 1]. GateNet was trained in a supervised manner through a weighted combination of the binary cross-entropy and soft-Dice loss functions on a dataset consisting of eventually 2336 images recorded in 7 distinct environments. The ground-truth mask for each sample in the training dataset was manually annotated. We augmented the training data through random affine transformations and variations in the HSV color space and artificial motion blur. The blur consists of the convolution of a squared averaging filter of random size between 5 and 15 pixels and random orientation. For the deployment on NVIDIA’s Jetson Xavier, we ported the network to TensorRT 5.0.2.6 with a batch size of 1 and full precision FP32 mode.

We deployed a different GateNet version in each competition race, with the only differences being the size of the training dataset and the data augmentation mechanisms. Networks were always trained from scratch when changing the augmentation mechanisms but otherwise fine-tuning previous models when expanding datasets. Before new races, we quickly fine-tuned the models to deployment environments with training data from the test sessions.

\subsubsection{Snake-gate: Active vision for corner identification}

To then retrieve the inner and outer corners of only the next gate, even when multiple gates were in sight, we employed a variation of the lightweight, active-vision algorithm \cite{Bajcsy2018} known as snake-gate \cite{li2020autonomous}. This two-stage, iterative sampling method reports the desired gate corners in a fixed order. The first stage starts at the intersection of the vertical and horizontal histograms of $\mathbf{M}$, which is the pixel with the highest probability of belonging to the closest gate. From that point, it starts sampling white pixels in a fixed direction in the image space (i.e., top-left, top-right, bottom-right, or bottom-left) until the corresponding outer corner is found. Thereafter, the sampling direction changes until all corners have been identified. A pixel is considered to be a corner if the sampling method cannot progress in the specified direction. Once the outer contour was identified, we used the centroid of this set of corners as the starting point to identify the inner corners of the gate by sampling black pixels instead of white ones. To overcome incorrect corner association at bank angles greater than $|45|$ deg, the mask $\mathbf{M}$ was first de-rotated using the drone’s estimated roll angle around the optical axis. 

Snake-gate requires (i) the mask of a gate to be continuous, and (ii) no gate overlap in the image space. The first requirement was normally met thanks to our robust and accurate GateNet model (see \cref{fig2}). However, most of the AIRR tracks had gates placed in front of each other, violating the second requirement. To cope with this, we developed gate-prior.

\subsubsection{Gate-prior: Sanity check on the identified corner locations}
Snake-gate does not provide any form of confidence metric regarding the identified corners. Therefore we developed a sanity check on the identified corners. The expected 3D location of the next gate was projected into the image space, and is called ``gate-prior''. We then compared the sides and angles of both inner and outer contours of this projection to those of the identified gate and only accepted the validity of a corner if the error of the associated sides and angle was below 25\%. Rejected corners of a contour with at least two valid corners were corrected using the shape of the gate-prior (see \cref{fig2}B). This actively reduced the number of outliers and it improved the robustness to challenging scenarios that could lead to discontinuous masks (e.g., HDR scenes, fast motion, partial gate in the image) and gate overlap (see \cref{fig2}). If no valid corners were found at all for two full seconds, a recovery mechanism would override ``gate-prior''.

\subsubsection{Localization via perspective-n-point}

The size and approximate location and orientation of the AIRR gates were known in all races. The estimation of the drone’s position and orientation was found by solving the Perspective-n-Point (PnP) problem, using the up to eight identified corner locations in the image space and their corresponding 3D locations. As in \cite{li2020autonomous,li2020visual}, instead of relying on pure vision-based PnP, we combined it with the onboard attitude estimate of the drone to retrieve the camera’s 3D location, as this was shown to be more robust in drone racing conditions. We used the OpenCV library to solve the PnP problem, or more specifically: an iterative method based on Levenberg-Marquardt optimization, which minimizes the reprojection error and requires at least three point-correspondences.

\subsection{State estimation}

Attitude estimation was performed using a complementary filter fed by gyroscope and accelerometer data. Position and velocity estimates were propagated using a drag and thrust model in the ``flat-body'' frame ${}^{fb}R_\mathrm{W}$  shown in \cref{fig3}A, which is a local tangent plane rotated by the yaw $\psi$ of the drone. The predicted drag forces in the flat-body frame $(a_x^{fb},a_y^{fb})$ were modeled as: 

\begin{equation}
    \begin{bmatrix}
        a_x^{fb} \\
        a_y^{fb}
    \end{bmatrix} =
    \begin{bmatrix}
        \hat{d_x} & 0 \\
        0 & \hat{d_y}
    \end{bmatrix}
    \underbrace{
    \begin{bmatrix}
        c_{\psi} & s_{\psi} \\
        -s_{\psi} & c_{\psi}
    \end{bmatrix}}_{{}^{fb}R_\mathrm{W}}
    \begin{bmatrix}
        v_x^\mathrm{W} \\
        v_y^\mathrm{W}
    \end{bmatrix}
\end{equation}

where $c_{\psi}$ and $s_{\psi}$ present the cosine and sine of the yaw angle, $(v_x^\mathrm{W}, v_y^\mathrm{W})$ the velocities in the world frame and the linear drag parameters $\hat{d_x}$ and $\hat{d_y}$ were found by fitting the integrated path to best match the know gate locations using flight logs. To reduce the drift of drag-model predictions in the world frame $a_\mathbf{W}$, an additional first-order linear filter called ``alpha'' fused the drag force model with accelerometer measurements (subscript $m$). The resulting prediction model is:

\begin{equation}
    \mathbf{a}_\mathrm{W} = \alpha \cdot {{}^{\mathrm{W}}R_{fb}} \quad  \begin{bmatrix} a_x^{fb} \\ a_y^{fb} \\ 0 \end{bmatrix}
    + (1 - \alpha)  ^\mathrm{W}R_{B}
    \begin{bmatrix} {a}_x^{B} \\ {a}_y^{B} \\ 0 \end{bmatrix}_{m} + \begin{bmatrix} 0 \\ 0 \\ g \end{bmatrix} + ^\mathrm{W}R_{B} \begin{bmatrix} 0 \\ 0 \\ {a}^{B}_z \end{bmatrix}_{m}
\end{equation}

where $\alpha$ determines the ratio between predictions based on drag model or accelerometers, $^\mathrm{w}R_{B}$ is the rotation from body-to-world, $g$ is the local gravitational acceleration and $\mathbf{a}^{B}_x,y,z$ are the accelerometer measurements. A value of $\alpha$ = 85\% was found to yield the best predictions. The predicted velocity $v^w$ and position $p^w$ in the world frame were obtained through integration: $\mathbf{v}_{W} = \textstyle{\int{\mathbf{a}_{W}}}$  and $\mathbf{p}_{W} = \textstyle{\int{\mathbf{v}_{W}}}$. 

For the altitude, a Kalman filter merged the 6Hz cutoff low-pass filtered vertical accelerations and the 50Hz cutoff low-pass filtered attitude-corrected downward-facing laser range finder readings. Position corrections were performed by merging the PnP estimates in world coordinates from the perception pipeline. As vision estimates occasionally still contained large errors and could have significant jumps, a moving horizon estimator (MHE) based on random consensus (RANSAC) was used. It is directly adopted from \cite{li2020visual}, but besides position and velocity corrections, yaw corrections were also made to account for initial heading alignment errors and yaw integration drift. The corrections were done by running the MHE filter independently on each axis $(p_x,p_y,\psi)$. Separate buffers with a maximum of 180 samples hold information about PnP estimates and delay-compensated inertial estimates. Samples older than 2 seconds were removed. The delay was fixed to 20 measurements (0.04 seconds) on the drone and 110 when run as HiL (Hardware in the Loop) simulation. Using random consensus with 200 iterations with 80\% of the samples, the filter fitted the predicted world path and heading with the PnP measurements. The result was a position correction $\hat{e_{p_{x,y}}}$ and a velocity correction $\hat{e_{v_{x,y}}}$ on top of the predicted estimates to obtain the current state at each timestep. The heading correction $\hat{e_\psi}$ on the other hand, was only applied once upon passing each gate. The least-squares fit for RANSAC was written as $\hat{x} = (A^T A + \partial I)^{-1} A^T y$ where the prior $\delta$ ensured a preference for small corrections in velocity estimates, and $\hat{x}$, $A$ and $y$ were defined to map the position and velocity errors in function of time $\Delta t$ over the buffer with samples $n=1$ to $N$ (given only for $p_x$):

\begin{equation}
    \underbrace{
    \begin{bmatrix}
        \Delta p_x |_{n = 1} \\
        \Delta p_x |_{n = 2} \\
        \vdots \\
        \Delta p_x |_{n = N} \\
    \end{bmatrix}}_{y}
     = 
     \underbrace{
    \begin{bmatrix}
    1 & \Delta t |_{n=1} \\
    1 & \Delta t |_{n=2} \\
    \vdots & \vdots \\ 
    1 & \Delta t |_{n=N}
    \end{bmatrix}^{\dagger}}_{A}
    \cdot
    \underbrace{
    \begin{bmatrix}
        \hat{e}_{\mathrm{p}} \\
        \hat{e}_{\mathrm{v}}
    \end{bmatrix}
    }_{x}
\end{equation}

This estimator ran in a separate thread and was executed each time there were enough samples in the buffers. When a gate was crossed, the prediction was reset to the value of the state and the MHE buffers were cleared. A minimum number of 27 PnP estimates (18 in simulation) were then required before the solution was allowed to jump to the new estimate.

\subsection{Path planning}

Path planning was done by tracking position waypoints from a list of approximate gate locations provided by the organizers. The altitude setpoint was kept constant at 1.75 meters. To better align with gates, the current commanded position $p_{x,y,z(cmd)}$ was placed 6 meters perpendicularly before the gate along its so-called ``centerline.'' When the robot got closer than 7 meters to this target, the point remained at 7 meters from the drone and moved towards the gate along the centerline until reaching the gate center.

\subsection{Control}

The selected active camera was either the right-center camera for tracks with right turns or the left-center camera for tracks with only left turns. The heading $\psi$ was commanded to align the active camera with the next gate center. This maximized the time gates were in-view and minimized the open-loop odometry phases. When arriving close to a gate, heading commands could get unnecessarily aggressive and reduce the quality of the model-based predictions. The yaw rate was therefore limited to 180 deg/s and the robot even stopped aligning the camera with the current gate when getting closer than 2.2 meters from the gate. This corresponds to the point where the gate would not be completely visible anymore.

The PID position controller mapped the lateral position errors in the flat body frame $\Delta p_{x,y}^{fb}$  to commanded lateral velocities $v_{x,y}^{fb}$. An additional proportional term was be added to have to robot align with the gate sooner by computing the lateral distance towards the gate centerline $\Delta p_y^{gate}$. The total lateral control became:

\begin{equation}
 v_y^{fb} = ( 1- \alpha_{center}) \cdot k_{p1} \cdot \Delta p_y^{b} + \alpha_{center} \cdot k_{p2} \cdot \Delta p_y^{gate}
\end{equation}

where $k_{p1}=0.45$ and $k_{p2}=0.45$ were gains and the mixing parameter $\alpha_{center}$ would determine if the robot flew directly to the waypoint along the shortest path or followed the gate centerline to improve perception and improve approach angles at the cost of increased distance to fly. Since the distance between gates was small in the last races, no obstacles were present along the gate centerlines, and some gates were placed at shallow angles, in the end, a value of $\alpha_{center}$ = 60\% was used. 

The forward velocity was a function of the distance to the gate $\Delta p_x^{fb}$ and the current motion vector. Far from the gate ($>$10 meters), the winning entry used a commanded velocity of 7.5 m/s. Then the speed was reduced to an alignment speed of 5.5 m/s. Once the state estimation would predict that the robot was sufficiently well aligned to pass through the gate within 80 centimeters of the center, it was allowed to speed up as much as possible. If the robot got so close to the gate that the gate was not in-sight anymore, to minimize the open-loop time spent in the gate it would always accelerate if it had not reached at least the gate-crossing speed of 7.5 m/s.

A velocity control loop converted the velocity commands to desired pitch and roll angles using a feed-forward gain of 0.009 rad/m/s and a velocity error feedback gain of 0.4 rad/m/s. The commanded pitch angle was constrained between -45 and -14${}^\circ$ pitch down, hence preventing pitching up. This served in keeping a good forward speed and helped perception as the fixed upward-looking angle of the camera meant that the bottom of the gate could fall outside the field of view when pitching up. Moreover, slower speeds and fast decelerations into the own propeller downwash also led to a larger drift of our drag-based odometry approach. The total bank angle was saturated at 45 degrees by maintaining the ratio between pitch and roll and is referred to as coupled saturation (S.2 in \cref{fig4}D). Finally, a rate limiter of 320 deg/sec was applied to reduce the effects of attitude changes on the available throttle.

Thrust commands were generated using traditional PID with a feedforward hover-thrust of 67\% at sea-level and 73\% in Littleton, which scaled with the inverse cosine of the total bank angle of the drone. Saturations were applied to the altitude error $\Delta p_z^w$ (+/- 2 meter) and  $T_{cmd}$ (15 - 100\%). An integrator windup protection was added to the PID loop by not integrating when the $T_{cmd}$ saturation was active. 

When the throttle would saturate in full throttle, a ``pitch-for-altitude'' controller was activated. As the throttle saturation could occur both in forward flight as well as in turns, instead of implementing a traditional ``pitch-up to climb'' controller, a max-bank reduction was used on top. In fast forward flight, the pitch-for-altitude controller would command pure pitch-up while during saturating turns the maximum roll angle of 45 degrees of roll was also reduced by the same amount.

Attitude control was achieved by computing feed-forward rate commands for the BetaFlight low-level autopilot that was running a pre-tuned rate controller. This was augmented with a bounded feedback controller on the error between the commanded and the current attitude. The errors in attitude are given as $e_\phi,e_\theta,e_\psi$ while the feedback and feedforward gains are $k_p=0.12$ and $k_{ff}=1/dt$. The change in desired pitch and roll angles in the given discrete timestep are noted $\Delta\theta_{cmd}$ and $\Delta\phi_{cmd}$ with timestep $dt$. The rate commands in roll, pitch, and yaw $p_{cmd},q_{cmd},r_{cmd}$ then become:

\begin{equation}
    \begin{bmatrix}
    p_{cmd} \\ - q_{cmd} \\ r_{cmd}
    \end{bmatrix} =
    \begin{bmatrix}
    1 & 0 & -s_{\theta} \\
    0 & c_{\phi} & c_{\theta} s_{\phi} \\
    0 & - s_{\phi} & c_{\theta} c_{\phi}
    \end{bmatrix}
    \begin{bmatrix}
    k_{ff} \Delta\phi_{cmd} + k_p e_\phi\\ 
    f_{ff} \Delta\theta_{cmd} + k_p e_\theta \\ 
    k_p e_\psi
    \end{bmatrix}
\end{equation}

These were scaled and sent together with the commanded thrust to the Betaflight low-level controller at 50 Hz. The low-level controller was the same for all teams and could not be altered.

All threads had sufficient spare processing power to log all variables described above, including all images and GateNet results. The gains were tuned based on a total of 60 short remote outsourced flight tests, lasting 5 to 15 seconds, after which logs would be returned. The test flights were performed at a separate roughly 60\% smaller track with a different density altitude than at the competition locations. This altered the drone dynamics, made flights very short, limited the types of maneuvers, and made it hard to reach full speed. The parameters were then manually fine tuned during the 1 hour test slot the day before the races (See time line \cref{fig1}B). 

\section{Analysis and Results}

In this section, we show the impact of the various elements of our approach on its performance, for perception, state estimation, and control. These experiments are conducted with a combination of real data collected with the drone, and synthetic data from the hardware-in-the-loop simulation platform provided by the competition organizers. Regarding perception, we assess the accuracy and robustness of the different GateNet models that we used throughout the competition qualitatively and quantitatively. Additionally, we provide an overall view of the computational expenses of the perception pipeline. Subsequently, we compare the performance of the developed state estimation scheme with that used in previous competitions. Concerning our control strategy, we conduct a detailed investigation of the various control improvements we introduced. Next, we determine the robustness of our approach to inaccuracies in the coarse drone race map. Lastly, we discuss the results of the competition, for the qualification stage, the seasonal races, and the final winning championship race.

\subsection{Perception}

\begin{figure}[hbt!]
    \centering
    \includegraphics[width=\textwidth]{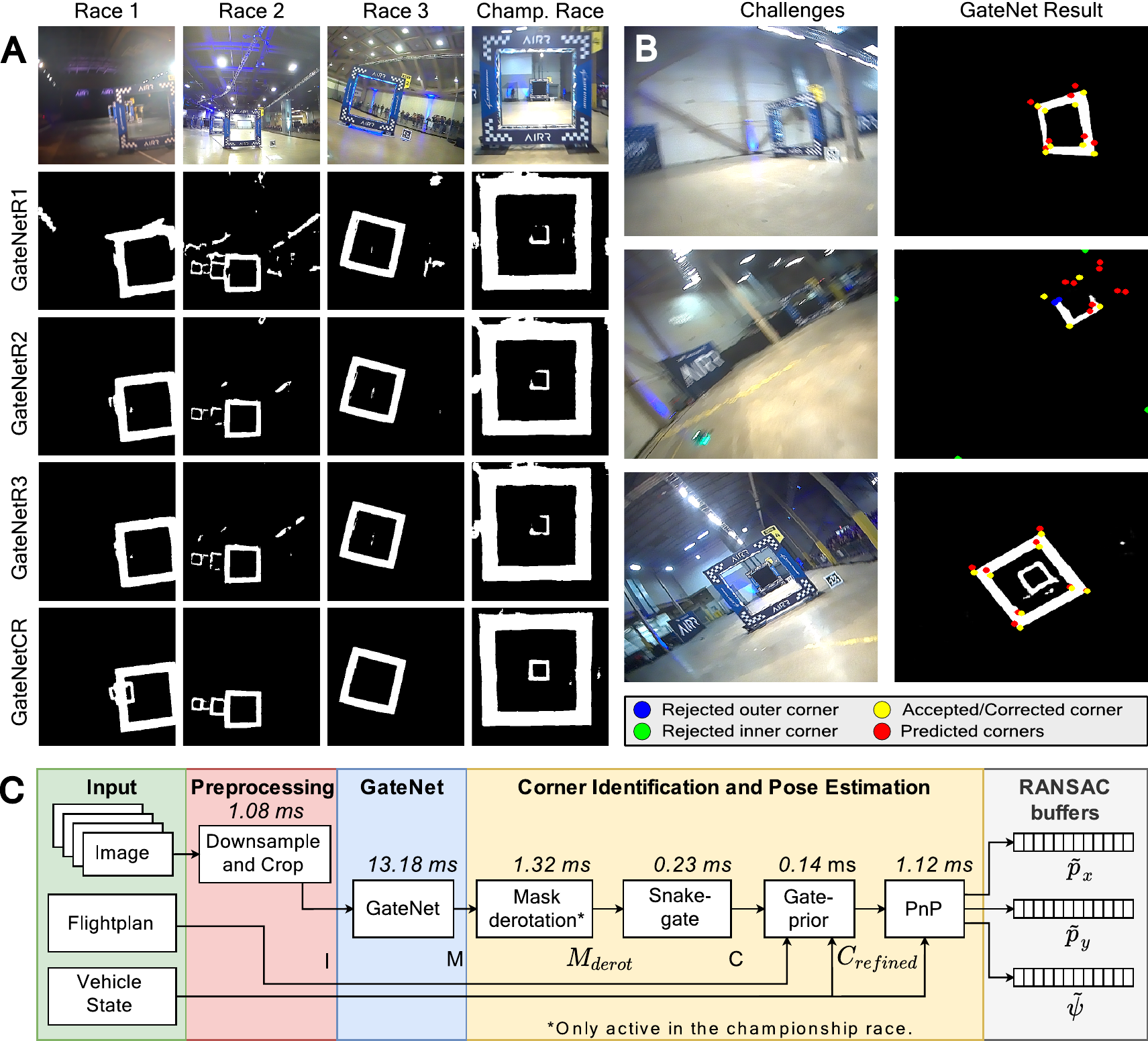}
    \caption{\textbf{Qualitative results of our perception pipeline.} \textbf{(A)} Segmentation masks generated by the four different GateNet models deployed throughout the competition from onboard images of each of the events. \textbf{(B)} Robustness of our GateNetCR-based perception pipeline to motion blur (top), distant gates leading to incomplete segmentation masks (center), and gate overlap, and agile bank angles (bottom). \textbf{(C)} Overview of the perception pipeline and execution time of each of the submodules on the Jetson Xavier.}
    \label{fig2}
\end{figure}

The perception pipeline consists of four steps. It starts by segmenting the images using a trained deep convolutional neural network called `GateNet'. It then finds the gate corners with an active vision method called `snake-gate'. The pipeline continues with corner validation based on projecting expected gate corners in the image plane which is further referred to as `gate-prior'. Lastly, the identified corners are used to solve the perspective-n-point problem for pose estimation (\cref{fig2}C). To quantitatively compare the different GateNet models that we used in each of the AIRR races, we collected and manually annotated a dataset consisting of 165 images logged during our 12-second winning run in the championship race. This dataset is characterized by (i) motion blur on the images due to the high-speed profile achieved in this run, (ii) strong illumination changes, and (iii) a challenging environment with banners containing visual features similar to those of the gates along the course of the track. The reason for only using logged images from this race in this evaluation is that it is the only data that was not used in any training dataset. When tested on this dataset, the different GateNet models deployed in the three seasonal races (i.e., GateNetR1, GateNetR2, and GateNetR3) achieve an average intersection over union (IoU) of 70\%, 74\%, and 76\%, respectively. GateNetCR, the actual model deployed for the championship race achieves an average IoU of 87\%. Qualitative results in \cref{fig2}A also exemplify GateNet’s progressive improvement throughout the competition using images from each of the events.  The significant improvement from GateNetR3 to GateNetCR can be attributed to two factors. First, as for every race, we added more manually annotated images from the race environment, which in this case was important for learning to ignore the newly placed banners. Second, after the third race, we added an artificial motion blur data augmentation mechanism to the training pipeline. Note that this artificial blur was not applied to the ground truth masks to still promote sharp segmentations (see \cref{fig2}B, top).

The example images and segmentations in \cref{fig2}B show the performance of GateNetCR in challenging scenarios like motion blur, distant gates, and adverse lighting conditions. They also show the robustness of our corner association algorithm. Our computationally efficient gate corner detector called ``snake-gate'' identifies the inner and outer corners of the front panel of the gates by actively sampling a small percentage of the pixels of the segmentation result. Then, our state-prediction-based sanity check and refinement of identified corner locations called ``gate-prior,'' compares the sides and angles of the inner and outer contours of both the detected and expected gate in the image plane to neglect distractor gates when multiple gates are in sight (like overhead projected gates from the live video stream). It also corrects the location of the estimated corners in case snake-gate didn’t identify a corner properly. The resulting corrected corners are finally fed to PnP-based pose estimation. This allows the drone to localize itself with respect to the next gate even in the case of a discontinuous GateNet mask (see \cref{fig2}B, center) or gate overlap in the image space (see \cref{fig2}B, bottom). While the corner correction using predicted gate locations can introduce errors due to the drift of the drone’s odometry, it significantly reduces the number of outliers that are introduced into the state-estimation pipeline. An automatic reset mechanism was put into place in case all corners were rejected for too long due to drift.

GateNet models requires on average 13.18 ms and thus can be performed much faster than the camera update rate (i.e., 60 Hz). The estimated gate mask from the cropped and downsized input image is de-rotated using the estimated camera roll angle around the optical axis to prevent incorrect corner association and requires an average of 1.32 ms computing time. The active-vision-based snake-gate method requires accessing the intensity information of only 1.64\% of the pixels of a 360x360 mask, which translates to a workload of 0.23 ms per image. We used the full horizontal and vertical histograms of the masks for snake-gate initialization even though computationally more efficient alternatives exist \cite{li2020autonomous}\cite{li2020visual}. Gate-prior takes on average 0.14 ms to correct the identified corners, and lastly, solving the perspective-n-point to localize the drone with respect to the gate requires 1.12 ms. Overall, the perception pipeline takes 17.07 ms while the thread runs at an average of 54 Hz. 

\begin{figure}[hbt!]
    \centering
    \includegraphics[width=\textwidth]{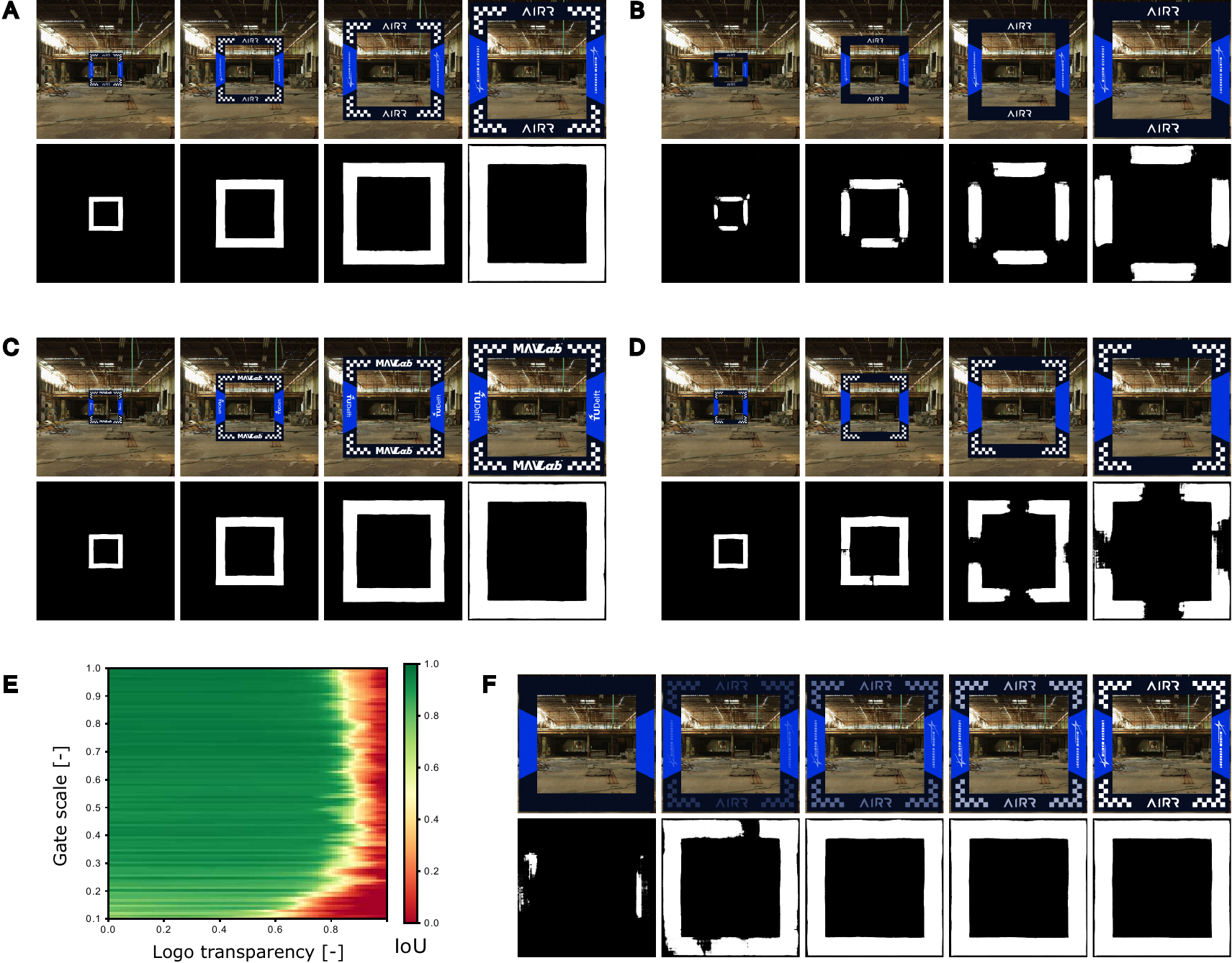}
    \caption{\textbf{How does GateNet perform gate segmentation? Qualitative and quantitative results of the impact of the gate's features on the segmentation accuracy.}
Different manipulation techniques were employed: \textbf{(A)} variation of the scale of the gate in the image space, \textbf{(B)} removal of checkerboards, \textbf{(C)} substitution of default logos with our own, \textbf{(D)} removal of logos, and \textbf{(F)} variation of feature transparency. \textbf{(E)} shows Intersection over Union (IoU) as a function of the scale of the gate and the feature transparency. In all cases, the maximum scale of the gate was set so that the four outer corners of the gate coincide with the extremes of the image space, of size (360, 360).}
    \label{figS1}
\end{figure}

Due to the importance of GateNet in the perception pipeline of our approach, we are also interested in understanding the impact that each of the features on the AIRR gates has on the segmentation accuracy. To this end, we conducted an experiment with synthetic data in which we varied the appearance of the gates’ features (i.e., checkerboards and text) and assessed the quality of the segmentation both qualitatively and quantitatively. As shown in \cref{figS1}, the appearance of the gate was manipulated in multiple ways by (i) changing the scale of the gate in the image space, (ii) removing the checkerboards, (iii) substituting the default logos with our own, (iv) removing the logos, and (v) modifying the transparency of both logos and checkerboards. 
Results in this figure confirm the importance of the contrast changes introduced by the logos and checkerboards. Removing the logos or checkerboard patterns lead to local gaps in the segmentation (\cref{figS1}B\&D). The test with the different logos confirms that GateNet has not learned the specific shapes of the AIRR logos but exploits more generic contrast in this region. Indeed, swapping the logos does not affect the segmentation (\cref{figS1}A\&C). The transparency test shows the importance of the presence of contrast, but also demonstrates that GateNet is quite robust to low contrasts (\cref{figS1}F). There is a dependency on the scale. \cref{figS1}E shows the intersection-over-union (IoU) for different gate scales and transparencies. At most scales, GateNet’s performance only breaks down at ~80\% transparency, whereas at the smallest scale (0.1) it breaks down at 60\%.

\subsection{State estimation}

\begin{figure}[hbtp]
    \centering
    \includegraphics[width=\textwidth]{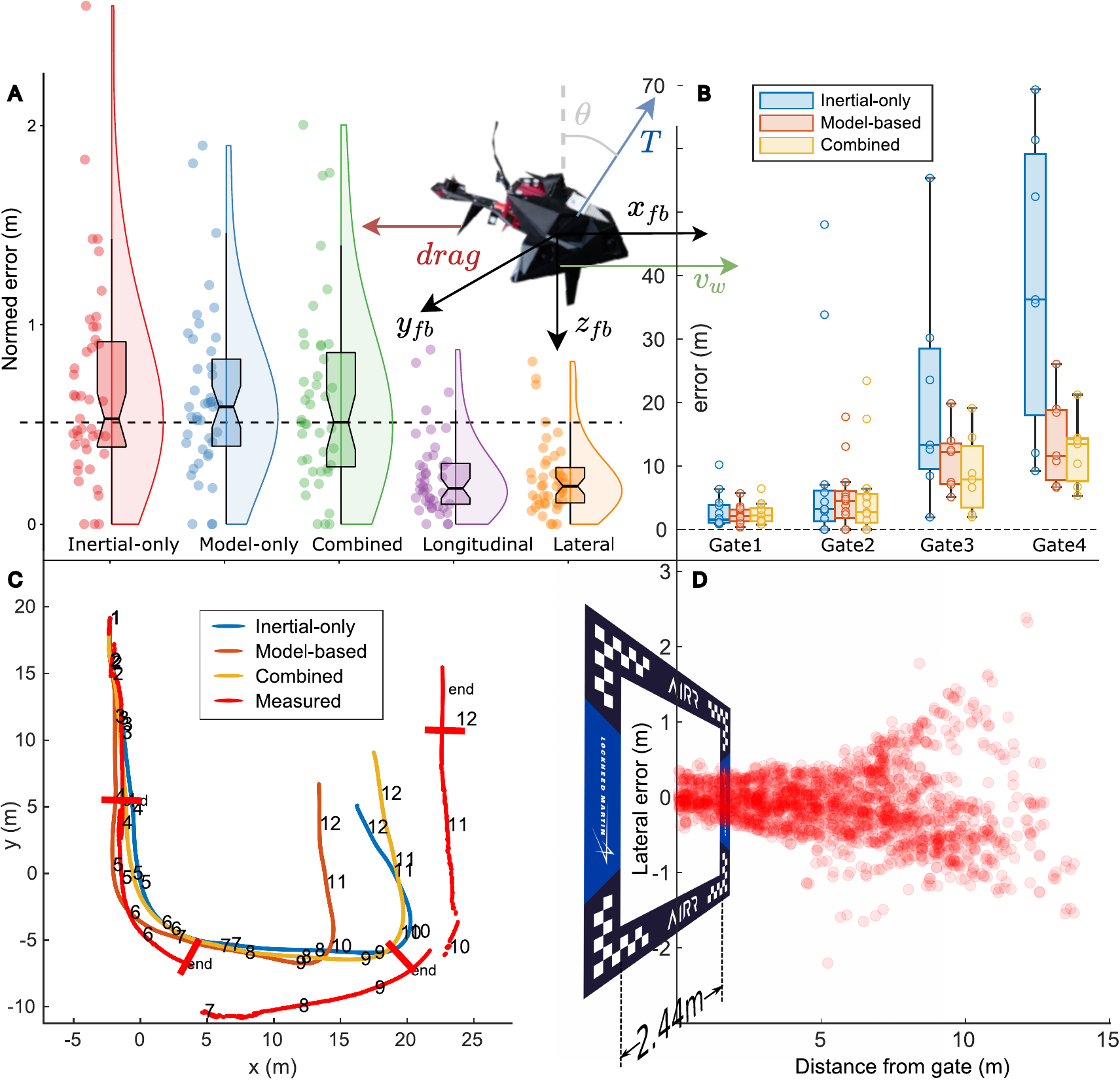}
    \caption{\textbf{Odometry results based on the real-world data from all competition and training runs (A, B, C) and state estimation error at the gate measured in simulation (D)}. \textbf{(A)} Average of errors in odometry of the three dynamic models (inertial-only, model-only ``flat-body'' and combined ``alpha'', and the lateral and longitudinal components of ``alpha'') after 1.8 seconds of prediction, where the gate size is indicated as a dashed line. \textbf{(B)} Statistics of the total accumulated odometry errors from the starting podium to each gate for the 13 full tracks flown competition tracks. \textbf{(C)}  Odometry-based position estimate top view for a typical track. Note that the third gate had a significantly different orientation than expected from the flight-plan, which causes the vision measurements to appear rotated. \textbf{(D)} Accuracy of state estimates as a function of distance to a gate measured from simulation runs.}
    \label{fig3}
\end{figure}

The vision-based position estimates are fused with model-inertial-based odometry to smoothen the measurements and overcome periods in which no gates are detected. This odometry is primarily based on a linear drag model of the quadrotor in a local tangent plane that follows the heading of the robot and is further referred to as the ``flat body'' frame (see \cref{fig3}A). The values of the linear drag were fitted with the scarce data from the real flights. Under low flight speed and constant altitude assumptions, this easy-to-identify model was shown to be a reasonable approximation \cite{li2020visual}. To improve the predictions during more aggressive maneuvers, instead of fitting a more complex model for which insufficient data was available, we chose to fuse accelerometer data in the odometry. The difference in performance was compared between the drag-only model called ``flat-body'', the combined model-inertial ``alpha'' method (named after its $\alpha$ parameter to set the relative importance of the drag-model versus accelerometer odometry), and traditional body-frame accelerometer-only odometry.

Since no position ground truth is available for the competition flights, the comparison is made with drone observations and track knowledge instead. This can in theory be subject to scaling and offset errors, but as long as the robot perception matches its predictions, they can successfully be merged, just like walking animals merge step-based odometry with visual observations without the need for a calibrated meter representation.

Position measurements close to the gate are very precise thanks to the very good geometry of the triangulation. Moreover, passing the gate is a crucial part of the race and relies on odometry only for the last few meters. We therefore first compared the odometry methods on 1.8-second stretches just before a gate (see \cref{fig3}A for the statistics and 3C for example stretches). Subsequently, we integrated the odometry methods from start to end on 13 full tracks and compared it with the end-gate in the relatively accurate gate map ($<$1m displacements). The results of this can be seen in \cref{fig3}B and a specific track in \cref{fig3}C. The results show that the model-inertial ``alpha'' method obtains the best results, which is why we used it in the final championship race. In general, the model-inertial-based odometry can obtain very good results given the scarce resources it requires (50\% within $<$15m endpoint error without calibration for a 12 seconds prediction horizon). Nevertheless, it only seems well-suited for tracks that have sufficient gates to perform position corrections.

\subsection{Control and path planning}

\begin{figure}[hbtp]
    \centering
    \includegraphics[width=\textwidth]{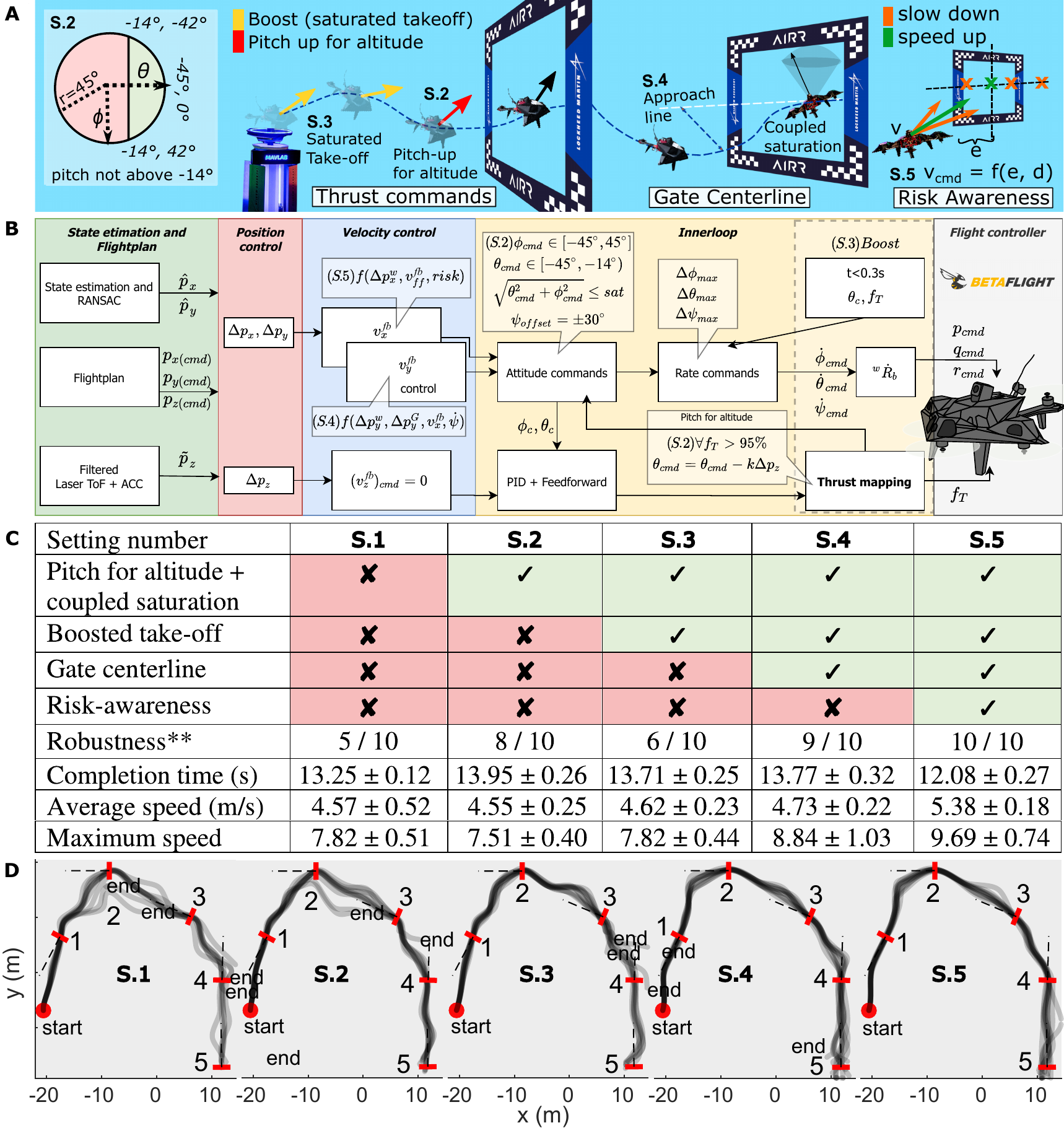}
    \caption{\textbf{Contribution of the various controller modifications in simulation on the Austin track.} \textbf{(A)} Illustration of the various controller modifications S.2 – S.5. \textbf{(B)} Schematic representation of the cascaded PID control pipeline but with several enhancements for fast and risk-aware flight. \textbf{(C)} Simulation test results for various configurations. \textbf{(D)} Top view of the tracks of the different controllers.}
    \label{fig4}
\end{figure}

To qualitatively validate the contribution to speed and reliability of the different control additions, a simulation study was performed. The initial classical control setup, marked as S.1 and shown in \cref{fig4}, only finished the track half the time when configured at competitive speeds. Four modifications were made to increase its speed and reliability.

The first modification to the classical control scheme combines the maximum roll and pitch angles into a single maximum bank called ``coupled saturation'' since separate maximum pitch and roll angles could yield 42\% higher bank angles when saturating together. Moreover, instead of putting a low and safe maximum bank angle of about 35$^\circ$ to never saturate thrust, we increased the limit past this point to 45$^\circ$. This would occasionally lead to insufficient thrust, which was addressed by introducing a pitch-for-altitude control loop. As shown in \cref{fig4} S.2, the pitch-for-altitude control loop leads to higher robustness, now finishing 8 / 10 runs with only minor speed loss.

An open-loop take-off with 100\% thrust and a saturating nose-down attitude called ``boost'' (S.3) was added that took off before the slow laser-range altitude sensor had finished booting. Likewise, a saturating pitch down was applied just before the final gate to get an even quicker finish in case the drone sensed it was well aligned. This reduces robustness (6 out of 10 runs finishing the track) but leads to slightly quicker flight times in simulation and much quicker finish times in real races by skipping the up to 1.5-second laser-range startup delay (not present in simulation).

Instead of moving along the shortest path towards the gate, an optimal approach line called ``gate-centerline'' was defined to prevent steep approach angles to gates. Too steep angles not only significantly decrease the safety margin of passing through gates but also affect the position dilution of precision (PDOP) of PnP corners, in turn reducing the quality of state estimates. This addition (S.4) increased the robustness to a success rate of 9 out of 10 through safer gate crossing angles and increased the quality of perception.

The final addition to the pipeline is an adaptive, risk-based longitudinal velocity control (\cref{fig4}, S.5).  At large distances from the gate, the drone is allowed to accelerate very significantly until it arrives at the optimal gate viewing distance, where it has to make sure the camera sees the gate. Once the drone is sufficiently confident that it is aligned properly with the gate, it can accelerate. On the other hand, when a gate is not at the expected location or takes longer to identify, or if the control fails to align quickly, the drone slows down. This combination of risk and perception awareness was simple to implement and very light, was intuitive to tune, and resulted in robust fast behavior. \cref{fig4}B S.5 shows that including risk-aware accelerations led to 10 successes out of 10 runs, while substantially increasing the average speed from 4.7 m/s to 5.4 m/s.

\subsection{Robustness}

Robustness was required to deal with the inability to assess calibrations of cameras and drones by the teams, random initial starting podiums, uncertainty about which drone was used for which race, and the inability to measure the track precisely (initially gate locations were even planned to change between runs). 

\subsubsection{Variations in the track}

\begin{figure}[hbtp]
    \centering
    \includegraphics[width=\textwidth]{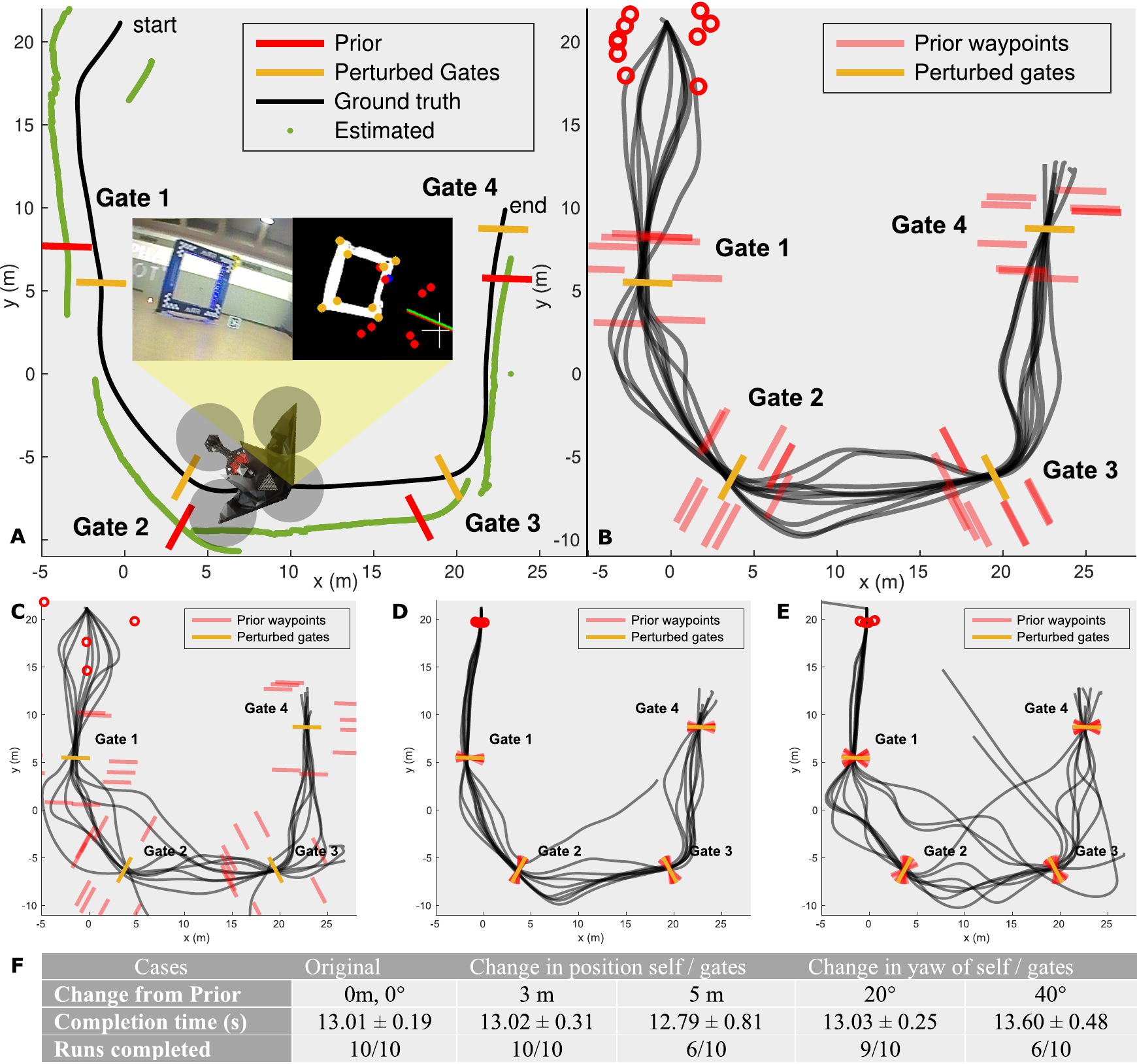}
    \caption{\textbf{Robustness to errors in internal track representation in simulation.} \textbf{(A)} The onboard state estimation based on the internal model (red gates) jumps to the correct solution after gathering sufficient evidence. \textbf{(B)} Ground truth position estimates of simulated runs on the Baltimore track with uniformly distributed 3m errors in flight-plan (red gates). Perturbations from smooth splined trajectories can be seen when the drone aligns with expected positions of gates, but quickly has to correct after observing the real position. \textbf{(C)} Uniform position offset of 5 m . \textbf{(D)} Uniform yaw perturbation of 20 degrees. \textbf{(E)} Uniform yaw perturbation of 40 degrees. \textbf{(F)} Overview of the perturbed flight-plans and the effect on lap completion times.}
    \label{fig5}
\end{figure}

To evaluate the robustness of our approach to changes in the racing environment, we performed a set of simulation experiments in which we perturb our drone’s internal representation of the individual gates and starting podiums in position and orientation. \cref{fig5}A shows that the state estimates quickly jump to the correct relative location with respect to the internal gate locations.  The red gates represent the deliberately biased gate locations in the robot’s internal map while the yellow gates mark the actual locations. The black line is the ground-truth drone trajectory, while the green line represents the drone’s internal state estimates. The advantage of this approach in a race with only a single lap is that the drone does not need to distinguish between its state error and internal map error. \cref{fig5}B shows that our pipeline can finish the course even with 3m perturbations in the course map, albeit with lateral swings due to the control initially aligning with a wrong gate location and then needing to correct. \cref{fig5}C,D,E show perturbations of 5m, 20$^\circ$ and 40$^\circ$. The table in \cref{fig5}F summarizes the robustness of the approach for various perturbations. The success rate only starts to drop substantially (to 6/10 runs) when gates are displaced 5m on a track where the inter-gate distance is about 12m. 

\subsubsection{Influence of camera calibration}

The AIRR competition was unique in that robots would only fly a single lap in a single heat per race, and the code only booted on the hardware for the first time at the second the race started. Although teams could ask which robot would be used in which heat, over the championship due to logistical reasons this could change, for instance, when robots were damaged by other teams before take-off or when insufficient robots were available to race against the human pilot and robots had to be re-used after falling into the final gate. This posed unique challenges to the developed AI to adapt to the unknown drone it was running on.

\begin{figure}[hbtp!]
    \centering
    \includegraphics[width=\textwidth]{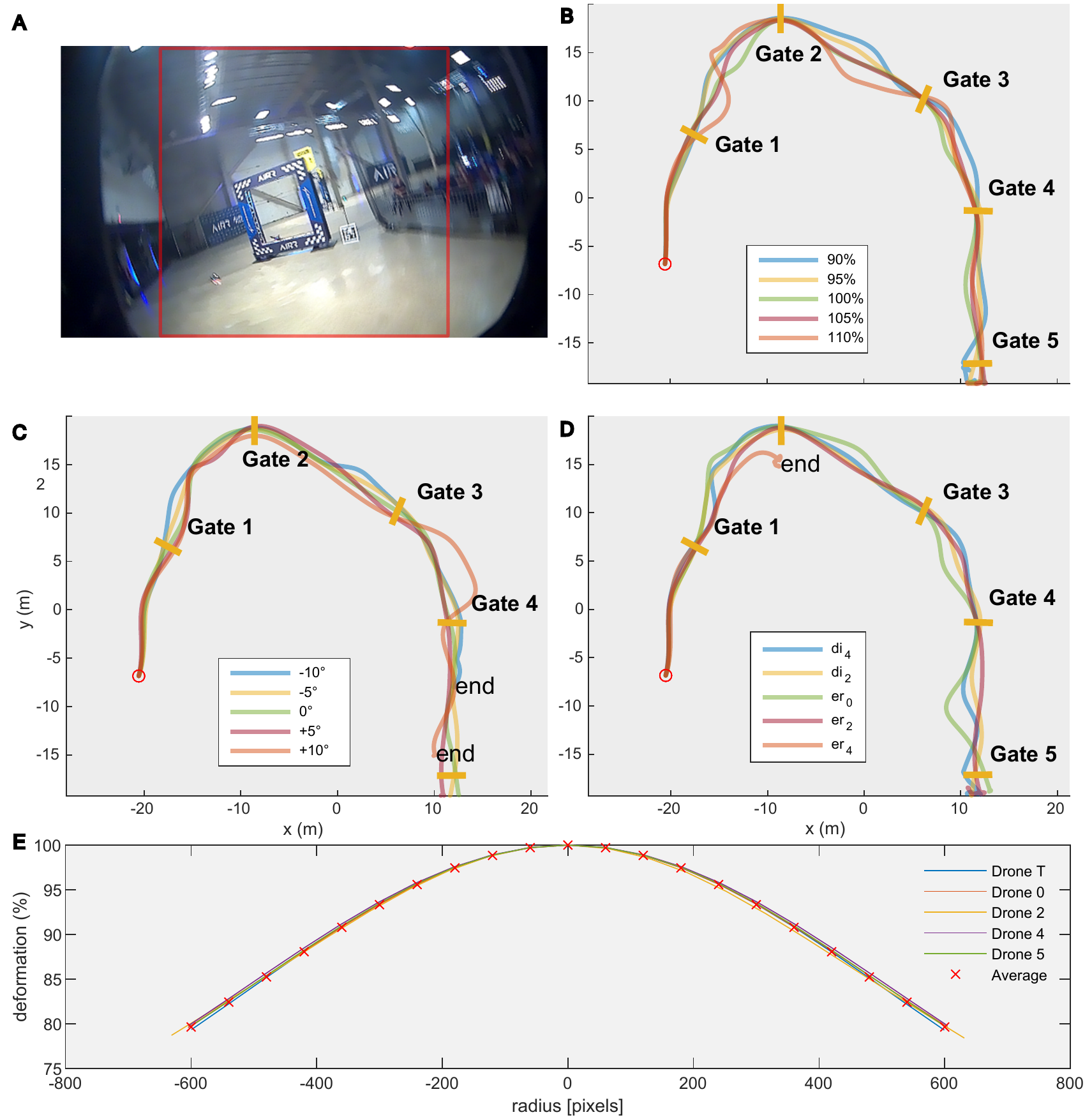}
    \caption{\textbf{The effect of camera calibration.} \textbf{(A)} RAW image from the Racer-AI showing the black borders of the camera protection cap and the cropped area in red, \textbf{(B)} Simulation results of flying the Austin competition track with an erroneous focal length, varied from 90\% to 110\% of the correct value. \textbf{(C)} Simulation results of flights with erroneous camera alignment value. \textbf{(D)} Simulation results of the track with a 2 and 4-pixel eroding or 2 and 4-pixel dilating filter applied to the GateNet mask. \textbf{(E)} Radial lens distortion as a function of pixel distance from the center point for individual calibrations of 5 different drones and the used average value.}
    \label{figS2}
\end{figure}

Furthermore, setting up the race drones was done within a very limited amount of time during a training session the day before the competition and was an important part of the challenge. Within this time slot, 60 minutes were available for flight tests with a single robot which, due to logistics, resulted in on average 2 to 7 short flights. This also meant that the other 4 robots were never flight-tested by the teams. The test data could then be analyzed in the remaining hours and used to prepare all the SD-cards with the final race code and calibrate all robots for the race the next day. The time required to calibrate the drones was considerable. Moreover, there was a risk that the calibration file for a specific robot was still not correct. This could happen after swapping hardware or after a crash that involved a camera repair. Hence, the choice was made to fly with a fixed hard-coded calibration for all robots and maximize the resilience of our algorithms against changes in calibration parameters.

The variation in lens distortion values within the given drone pool was analyzed (See \cref{figS2}) and found not to vary more than a few percent. Moreover, thanks to the modified PnP which used the attitude to better condition the triangulation, the variation in distortion caused minimal errors in the PnP position estimates that used the undistortion parameters. The figure shows the fixed ``average calibration'' used in all drones. Furthermore, the influence of errors in extrinsic parameters was studied in simulation. While better calibrations lead to better results, simulation results show that camera alignment errors of 10 degrees in the heading direction would still result in stable flight finishing the course. A change in focal length up to 10 percent was tested in simulation and had a direct influence on the scale at which the robot perceived the world and thus at which speed it flew but even in this case the robot still managed to finish the track. Finally, simulations were made to assess the influence of improper segmentation of the gate contours by GateNet. The masks were dilated and eroded with up to 4 pixels. This introduces dramatic distance varying scaling errors and inconsistencies between inner and outer gate corners. The erosion did also lead to gates being seen much later, and the robot failed to finish the race in the case of 4-pixel erosion of the mask. In all other cases, the robot still reached the final gate, although significant decreases in performance could be seen. This insensitivity to exact calibration played an important role over the season.

\subsection{Competition outcome}

\begin{figure}[hbtp]
    \centering
    \includegraphics[width=\textwidth]{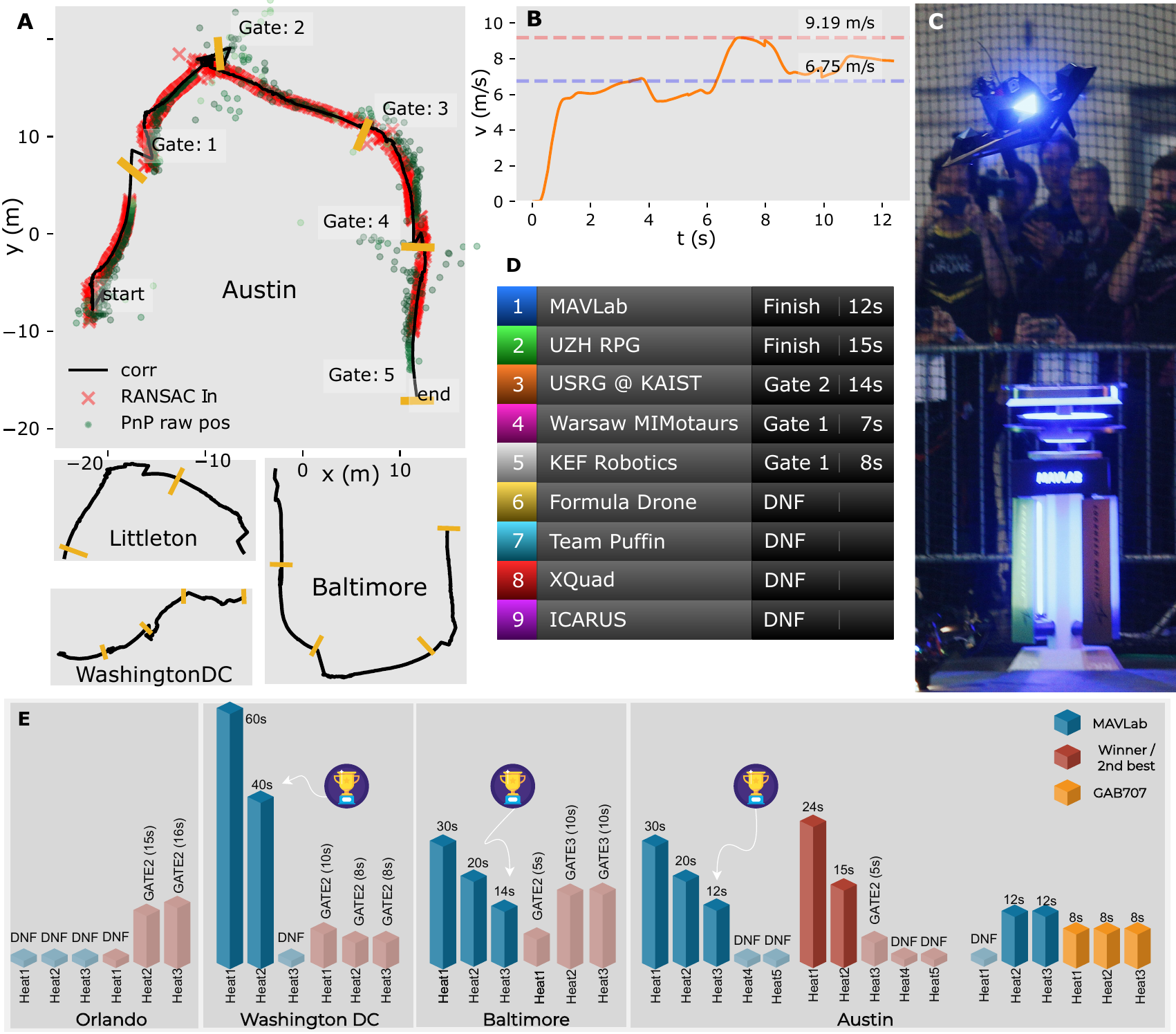}
    \caption{\textbf{Overview of our performance through the 2019 AIRR competition.} \textbf{(A)} Top view of the estimated executed path in relation to the previously-known rough map of each track. \textbf{(B)} Estimated speed profile at the championship race. Our drone flew with an average speed of 6.75 m/s and reached a top speed of 9.19 m/s. \textbf{(C)} Our MAVLab autonomous drone taking off at the championship race, before finishing the 74-meter course in 12 seconds (Picture credit: DRL) \textbf{(D)} Leaderboard of the championship race, indicating the time it took for each team to reach their farthest waypoint on the track. (DNF = ``did not finish''). \textbf{(E)} Completion times at the different tracks.}
    \label{fig6}
\end{figure}

\begin{figure}[phbt]
    \centering
    \includegraphics[width=\textwidth]{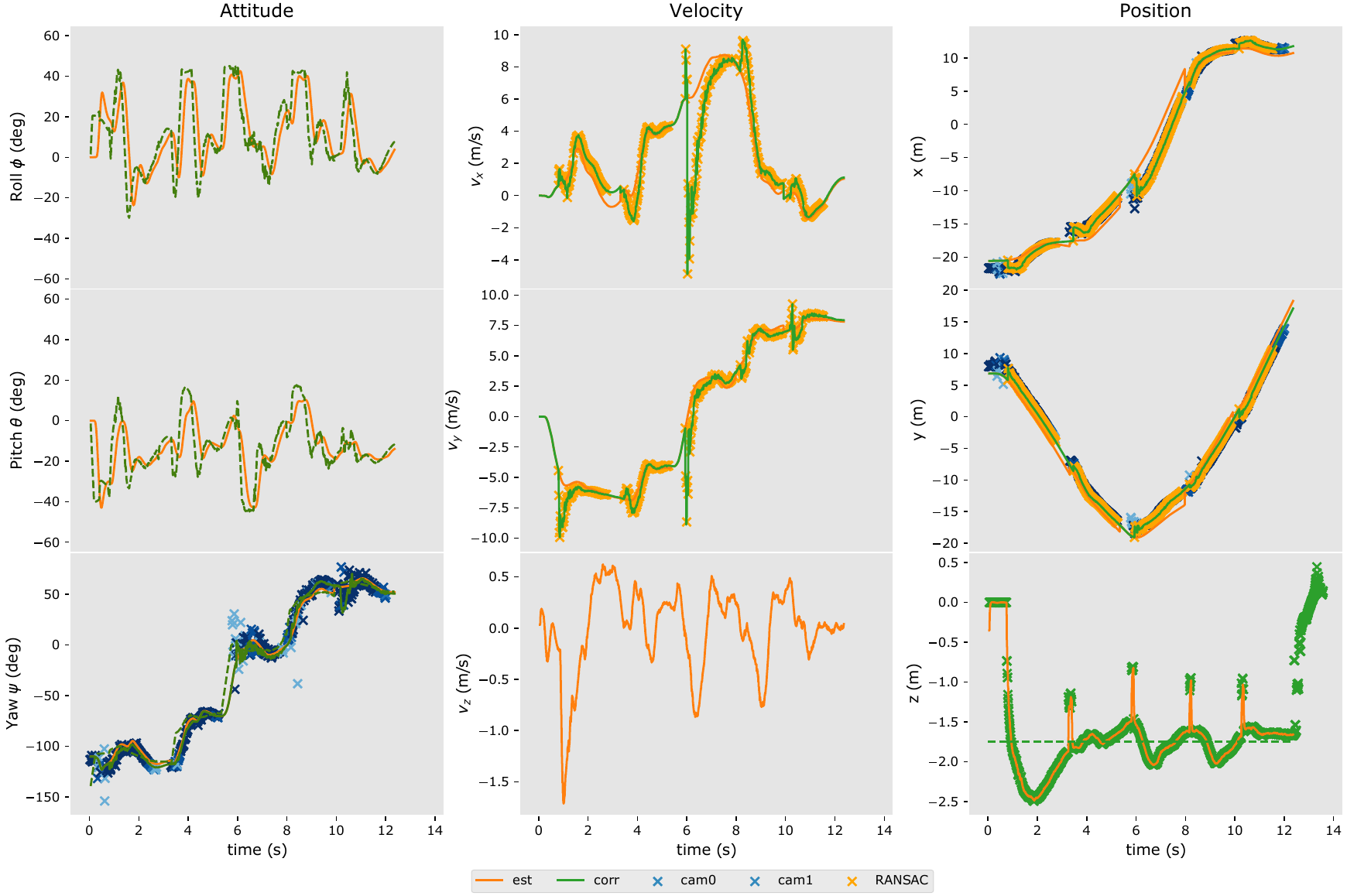}
    \caption{\textbf{Logs from the winning flight in the Austin race}. \textbf{Left column}: Attitude commands and estimates. \textbf{Middle column}: World frame velocity estimates. \textbf{Right column}: World frame position estimates.}
    \label{figS4}
\end{figure}

\begin{figure}[hbpt]
    \centering
    \includegraphics[width=0.85\textwidth]{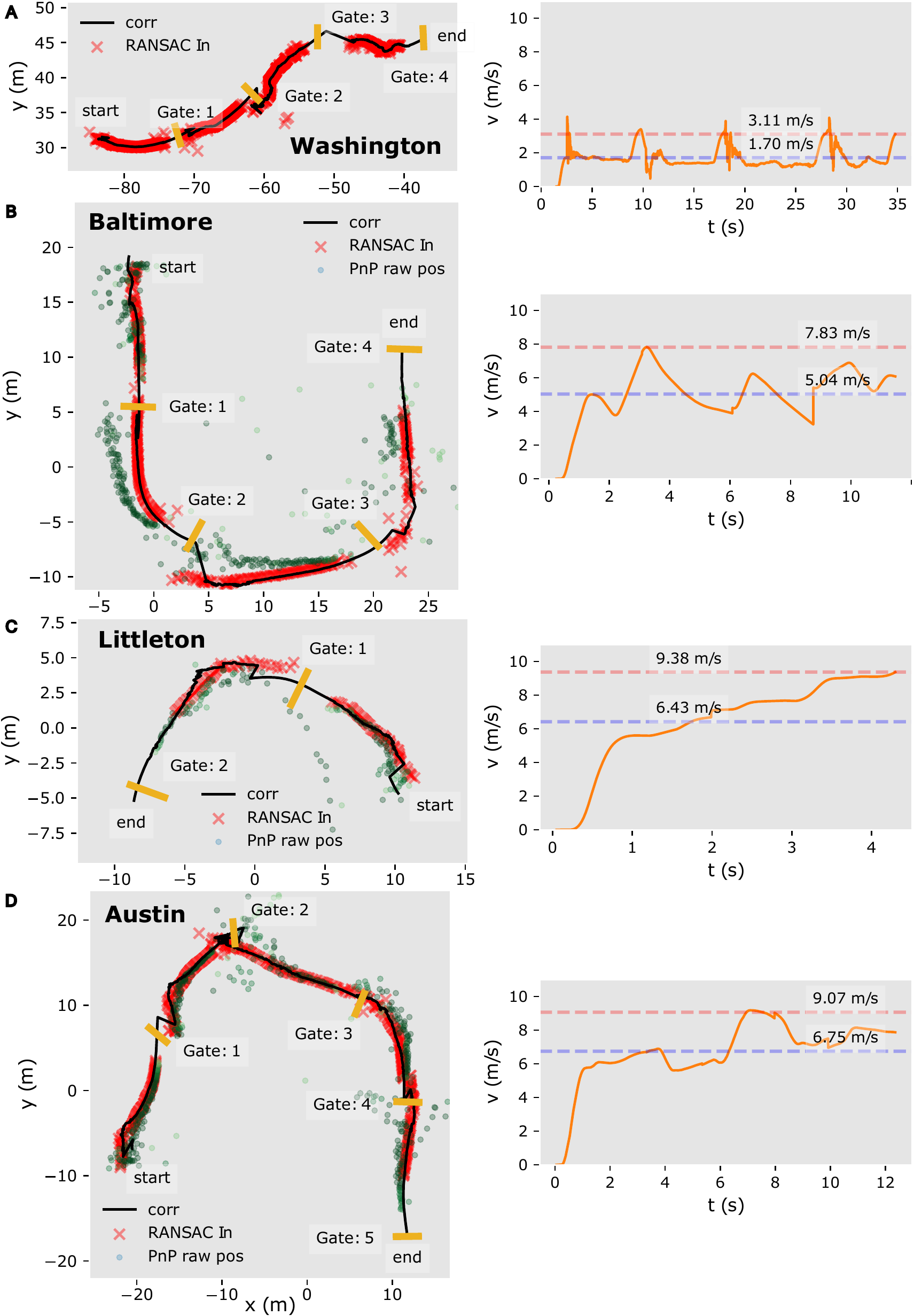}
    \caption{\textbf{Onboard state estimation of the position and velocity for the best flights of each race} of the AIRR season (arranged chronologically). Left column: World frame state estimates and trajectory of \textbf{(A)} Washington DC, \textbf{(B)} Baltimore, \textbf{(C)} Littleton, and \textbf{(D)} Austin. \textbf{Right column}: Respective world frame velocities.}
    \label{figS3}
\end{figure}

\begin{figure}[phbt]
    \centering
    \includegraphics[width=\textwidth]{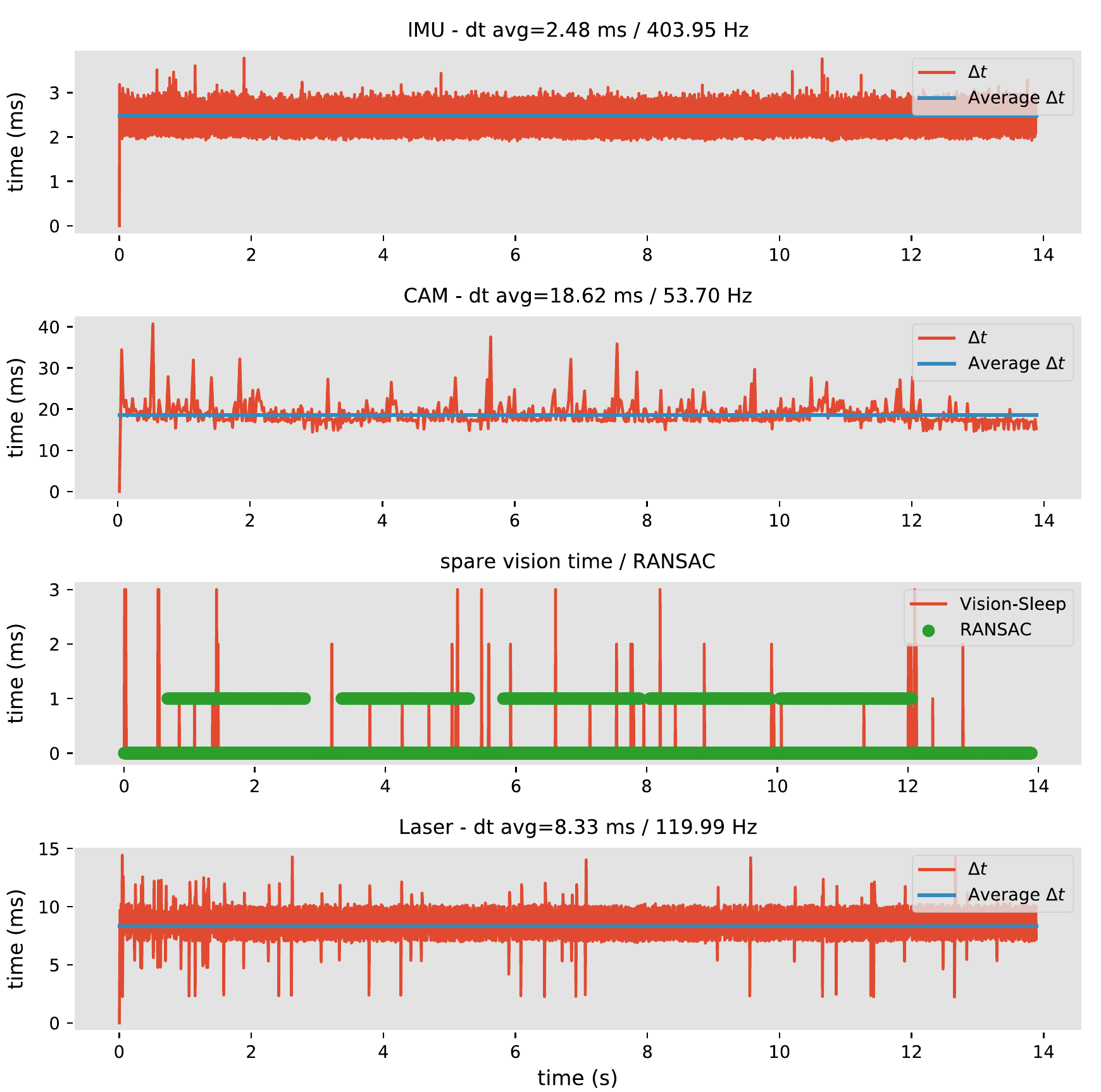}
    \caption{\textbf{Quality of threading}. Timestamps were collected from the threads responsible for gathering sensor data onboard the RacerAI. Their nature is observed to be non-deterministic but stays within bounds. Note that besides the sensor threads, the vision thread, the vision logging thread, the estimator thread, and the control thread were running in parallel and were data-driven.}
    \label{figS5}
\end{figure}

\cref{fig6} represents our competition results for the three seasonal races and the championship race and that of the best opponent. Each race consisted of several ``heats'' which could use a different version of the code. This allowed teams to ensure completion of the track with a steady speed in the initial heats and setting best finish times in the later heats. \cref{fig6}A shows the trajectories flown by our best run during all races while \cref{figS3}, \cref{figS4} and \cref{figS5} give extra details. Since there was no ground-truth position measurement system, we show the onboard position estimates (marked in green).

The first race of the season in Orlando was won by team KAIST from South Korea, who were able to fly through two gates of the track. Our drone was not able to pass any gate primarily due to unanticipated enormous differences in illumination between testing and competition. The second race took place in Washington DC and our team was the first to fully finish any track. Since then, our racing speed significantly increased over the seasonal races. The championship race held at Austin was a genuine race with tight competition as multiple teams were finishing the track during their training and qualification races. The best performance of the finalists is shown in \cref{fig6}D. Our finish time of 12 seconds with an average velocity of 6.75 m/s and maximum velocity of 9.19 m/s was the prize-winning run.

After winning the AI v/s AI challenge, our drone was staged against the DRL champion GAB707 in a Human v/s AI challenge. In this race where we were not allowed to change any parameter, our first deployment led to a crash into the first gate, due to a change in starting-position more than 50\% of the distance to the gate. Both other heats against the human also started from unanticipated podiums but were within the robustness of the system and finished with the same 12 second lap time.


An overview of the best tracks flown at the different locations is given in \cref{figS3} and Movie S2. Attitude, speed, and position data in function of time are given for the winning flight in \cref{figS4}. The timing of sensor data is depicted in \cref{figS5}. The logfiles of the winning flight of the Austin championship are given in \cite{AIRR_LOGS_2019}.

\section{Conclusion and Discussion}

We presented our approach to the AIRR competition, which led to winning two out of three seasonal races, the championship race, and led to the title of ``AIRR World Champion 2019''. Our approach was human-inspired in the sense that the developed AI focuses on the drone racing gates, which serve as waypoints for the race trajectory, relies significantly on model-based odometry, and accelerates as much as possible when the situation is safe. The approach successfully dealt with the scarcity of data and was highly computationally efficient, allowing for a very fast perception and action cycle. By having a deep-neural-network vision front end, our approach proved to be particularly resilient to frequent changes in the environment. The only occasion that the changes proved to be too much was during the first season race where all training was done in ambient light conditions while the race took place in showbiz illumination conditions that completely over-exposed the gates.

AI purists may raise the question of how much the competition, and our approach, was actually about AI. In a ``pure'' AI scenario, the drone’s perception and control would have been learned from scratch, making use of the provided simulator. Such an approach would, however, have clashed with the competition setup and timeline. The simulator was ready only a few weeks before the first race and had a substantial reality gap in terms of the drone’s dynamics and image capturing. For example, the images in the simulation had a variable delay, going up to 0.5s, which was worse than on the real platform. Combined with the extremely scarce access to the drone and outsourced testing, this would have left very little time for end-to-end training and a successful crossing of the reality gap.

We strongly believe though that robotics competitions like AIRR reveal highly relevant research areas for AI. In this case: \emph{How can AI best be designed, so that robots need minimal time and data to reach robust and highly agile flight?} A monolithic neural network trained end-to-end purely in simulation likely requires too many training samples to form the best answer to this question. And, if we equate the experience accumulated in a simulator with the evolutionary experience before the birth of an animal, this is not the strategy that we observe in animals either. Animals – even from the same species – are all different physically, and their intelligence is set up in such a way as to deal effectively with these differences. Whereas humans need a long development time before becoming operational, many flying insects can almost immediately fly and perform successful behaviors. The reason for this is that evolution has put in place various mechanisms to deal with, e.g., the physical differences between members of the same species, ranging from adaptation to various learning mechanisms. This means that true AI will require not only reinforcement learning \cite{sutton2018reinforcement}, but also, various types of self-supervised learning \cite{thrun2006stanley}, unsupervised learning \cite{kohonen1982self}, and lower-level adaptations as used for instance in adaptive control \cite{aastrom2013adaptive,Johnson2005}. This last level of learning, arguably at the lowest level, is hugely important for crossing the reality gap in robotics \cite{scheper2016abstraction}. To make our approach work on time and robustly enough for the competition, the employed AI still relied quite a lot on us as human system designers. We learned the drone’s model based on flight data, used supervised learning with human labeling of 2336 images for training GateNet, and designed an active vision algorithm for finding corners in the segmented images. For state estimation and control, we predominantly used human-engineered solutions that most would classify as being part of the field of control system theory and were adjusted by experts when the robot for instance moved to a location with very different air density.

Please note though that we and others are quickly developing deep learning approaches that can cross the reality gap for performing visual odometry \cite{Sanket2020}, tracking of predetermined optimal trajectories \cite{kaufmann2020deep} or even for full optimal control \cite{li2020aggressive}. AIRR has already been a driving force to develop AI methods that will successfully bridge the reality gap, even for robots that are difficult to model in detail upfront. If we succeed to do this for the heavily resource-constrained and time-pressed racing drones, then it will also generalize to other types of robots and tasks, such as autonomous vacuum cleaners and self-driving cars.


\subsubsection*{Acknowledgments}
We thank all the worldwide teams participating at the 2019 AIRR Circuit for providing collaboration and learning throughout the competition. A robotics AI challenge of this scale would not have been possible without the initiation of Lockheed Martin and The Drone Racing League. The organizing teams at LM and DRL worked tremendously hard, facilitating such a great outcome in the field of AI and Robotics. We were fortunate to have Jelle Westenberger, Anoosh Hegde, Sameera Sundaruwan, our entire drone race team since IROS 2016 back in Delft for helping us with the organization of the dataset and exploring high-level control strategies alongside their constant mentorship.

\subsubsection*{Author contributions}
All authors contributed equally to the writing of the manuscript. The GateNet and Gate-Prior were mainly developed by F. Parades with help during the training of the deep net from C. De Wagter. The control was mainly developed by N. Sheth with additions like gate approach line and risk-aware speed from C. De Wagter. The model-based state prediction was mainly developed by G. de Croon with the alpha fusion and quality analysis from N. Sheth. The state estimation, implementation, and writing were done equally by all authors. The team lead was G. de Croon and the competition lead was C. De Wagter.

\bibliographystyle{apalike}
\bibliography{airr_tudelft}

\begin{thebibliography}{}

\bibitem[{\AA}str{\"o}m and Wittenmark, 2013]{aastrom2013adaptive}
{\AA}str{\"o}m, K.~J. and Wittenmark, B. (2013).
\newblock {\em Adaptive control}.
\newblock Courier Corporation.

\bibitem[Bajcsy et~al., 2018]{Bajcsy2018}
Bajcsy, R., Aloimonos, Y., and Tsotsos, J.~K. (2018).
\newblock {Revisiting active perception}.
\newblock {\em Autonomous Robots}, 42(2):177--196.

\bibitem[Bangura and Mahony, 2014]{Bangura2014}
Bangura, M. and Mahony, R. (2014).
\newblock {Real-time Model Predictive Control for Quadrotors}.
\newblock {\em IFAC Proceedings Volumes}, 47(3):11773--11780.

\bibitem[Campbell et~al., 2002]{campbell2002deep}
Campbell, M., Hoane~Jr, A.~J., and Hsu, F.-h. (2002).
\newblock Deep blue.
\newblock {\em Artificial intelligence}, 134(1-2):57--83.

\bibitem[Cocoma-Ortega and Martinez-Carranza, 2019]{cocoma2019cnn}
Cocoma-Ortega, J.~A. and Martinez-Carranza, J. (2019).
\newblock A cnn based drone localisation approach for autonomous drone racing.
\newblock In {\em 11th International Micro Air Vehicle Competition and
  Conference}.

\bibitem[De~Wagter et~al., 2021]{AIRR_LOGS_2019}
De~Wagter, C., Paredes-Vallés, F., Sheth, N., and de~Croon, G. C. H.~E.
  (2021).
\newblock {Logfiles of The Artificial Intelligence behind the winning entry to
  the 2019 AI Robotic Racing Competition}.

\bibitem[Delmerico and Scaramuzza, 2018]{delmerico2018benchmark}
Delmerico, J. and Scaramuzza, D. (2018).
\newblock A benchmark comparison of monocular visual-inertial odometry
  algorithms for flying robots.
\newblock In {\em 2018 IEEE International Conference on Robotics and Automation
  (ICRA)}, pages 2502--2509. IEEE.

\bibitem[Falanga et~al., 2017]{falanga2017aggressive}
Falanga, D., Mueggler, E., Faessler, M., and Scaramuzza, D. (2017).
\newblock Aggressive quadrotor flight through narrow gaps with onboard sensing
  and computing using active vision.
\newblock In {\em 2017 IEEE international conference on robotics and automation
  (ICRA)}, pages 5774--5781. IEEE.

\bibitem[Floreano and Wood, 2015]{Floreano2015}
Floreano, D. and Wood, R.~J. (2015).
\newblock Science, technology and the future of small autonomous drones.
\newblock {\em Nature}, 521(7553):460--466.

\bibitem[Foehn et~al., 2020]{foehn2020alphapilot}
Foehn, P., Brescianini, D., Kaufmann, E., Cieslewski, T., Gehrig, M., Muglikar,
  M., and Scaramuzza, D. (2020).
\newblock Alphapilot: Autonomous drone racing.
\newblock {\em arXiv preprint arXiv:2005.12813}.

\bibitem[Jakobi et~al., 1995]{jakobi1995noise}
Jakobi, N., Husbands, P., and Harvey, I. (1995).
\newblock Noise and the reality gap: The use of simulation in evolutionary
  robotics.
\newblock In {\em European Conference on Artificial Life}, pages 704--720.
  Springer.

\bibitem[Johnson and Kannan, 2005]{Johnson2005}
Johnson, E.~N. and Kannan, S.~K. (2005).
\newblock {Adaptive Trajectory Control for Autonomous Helicopters}.
\newblock {\em Journal of Guidance, Control, and Dynamics}, 28(3):524--538.

\bibitem[Jung et~al., 2018]{jung2018perception}
Jung, S., Hwang, S., Shin, H., and Shim, D.~H. (2018).
\newblock Perception, guidance, and navigation for indoor autonomous drone
  racing using deep learning.
\newblock {\em IEEE Robotics and Automation Letters}, 3(3):2539--2544.

\bibitem[Kaufmann et~al., 2019]{kaufmann2019beauty}
Kaufmann, E., Gehrig, M., Foehn, P., Ranftl, R., Dosovitskiy, A., Koltun, V.,
  and Scaramuzza, D. (2019).
\newblock Beauty and the beast: Optimal methods meet learning for drone racing.
\newblock In {\em 2019 International Conference on Robotics and Automation
  (ICRA)}, pages 690--696. IEEE.

\bibitem[Kaufmann et~al., 2020]{kaufmann2020deep}
Kaufmann, E., Loquercio, A., Ranftl, R., M{\"u}ller, M., Koltun, V., and
  Scaramuzza, D. (2020).
\newblock Deep drone acrobatics.
\newblock {\em arXiv preprint arXiv:2006.05768}.

\bibitem[Kohonen, 1982]{kohonen1982self}
Kohonen, T. (1982).
\newblock Self-organized formation of topologically correct feature maps.
\newblock {\em Biological cybernetics}, 43(1):59--69.

\bibitem[Koos et~al., 2012]{koos2012transferability}
Koos, S., Mouret, J.-B., and Doncieux, S. (2012).
\newblock The transferability approach: Crossing the reality gap in
  evolutionary robotics.
\newblock {\em IEEE Transactions on Evolutionary Computation}, 17(1):122--145.

\bibitem[Krizhevsky et~al., 2012]{krizhevsky2012imagenet}
Krizhevsky, A., Sutskever, I., and Hinton, G.~E. (2012).
\newblock Imagenet classification with deep convolutional neural networks.
\newblock In {\em Advances in neural information processing systems}, pages
  1097--1105.

\bibitem[Li et~al., 2020a]{li2020autonomous}
Li, S., Ozo, M.~M., De~Wagter, C., and de~Croon, G.~C. (2020a).
\newblock Autonomous drone race: A computationally efficient vision-based
  navigation and control strategy.
\newblock {\em Robotics and Autonomous Systems}, 133:103621.

\bibitem[Li et~al., 2020b]{li2020aggressive}
Li, S., {\"O}zt{\"u}rk, E., De~Wagter, C., de~Croon, G.~C., and Izzo, D.
  (2020b).
\newblock Aggressive online control of a quadrotor via deep network
  representations of optimality principles.
\newblock In {\em 2020 IEEE International Conference on Robotics and Automation
  (ICRA)}, pages 6282--6287. IEEE.

\bibitem[Li et~al., 2020c]{li2020visual}
Li, S., van~der Horst, E., Duernay, P., De~Wagter, C., and de~Croon, G.~C.
  (2020c).
\newblock Visual model-predictive localization for computationally efficient
  autonomous racing of a 72-g drone.
\newblock {\em Journal of Field Robotics}.

\bibitem[Loianno et~al., 2016]{loianno2016estimation}
Loianno, G., Brunner, C., McGrath, G., and Kumar, V. (2016).
\newblock Estimation, control, and planning for aggressive flight with a small
  quadrotor with a single camera and imu.
\newblock {\em IEEE Robotics and Automation Letters}, 2(2):404--411.

\bibitem[Loquercio et~al., 2019]{loquercio2019deep}
Loquercio, A., Kaufmann, E., Ranftl, R., Dosovitskiy, A., Koltun, V., and
  Scaramuzza, D. (2019).
\newblock Deep drone racing: From simulation to reality with domain
  randomization.
\newblock {\em IEEE Transactions on Robotics}, 36(1):1--14.

\bibitem[Lupashin et~al., 2010]{lupashin2010simple}
Lupashin, S., Sch{\"o}llig, A., Sherback, M., and D'Andrea, R. (2010).
\newblock A simple learning strategy for high-speed quadrocopter multi-flips.
\newblock In {\em 2010 IEEE international conference on robotics and
  automation}, pages 1642--1648. IEEE.

\bibitem[Mellinger and Kumar, 2011]{mellinger2011minimum}
Mellinger, D. and Kumar, V. (2011).
\newblock Minimum snap trajectory generation and control for quadrotors.
\newblock In {\em 2011 IEEE international conference on robotics and
  automation}, pages 2520--2525. IEEE.

\bibitem[Mellinger et~al., 2012]{mellinger2012trajectory}
Mellinger, D., Michael, N., and Kumar, V. (2012).
\newblock Trajectory generation and control for precise aggressive maneuvers
  with quadrotors.
\newblock {\em The International Journal of Robotics Research}, 31(5):664--674.

\bibitem[Moon et~al., 2019]{moon2019challenges}
Moon, H., Martinez-Carranza, J., Cieslewski, T., Faessler, M., Falanga, D.,
  Simovic, A., Scaramuzza, D., Li, S., Ozo, M., De~Wagter, C., et~al. (2019).
\newblock Challenges and implemented technologies used in autonomous drone
  racing.
\newblock {\em Intelligent Service Robotics}, 12(2):137--148.

\bibitem[Moon et~al., 2017]{moon2017iros}
Moon, H., Sun, Y., Baltes, J., and Kim, S.~J. (2017).
\newblock The iros 2016 competitions [competitions].
\newblock {\em IEEE Robotics and Automation Magazine}, 24(1):20--29.

\bibitem[Morrell et~al., 2018]{morrell2018differential}
Morrell, B., Rigter, M., Merewether, G., Reid, R., Thakker, R., Tzanetos, T.,
  Rajur, V., and Chamitoff, G. (2018).
\newblock Differential flatness transformations for aggressive quadrotor
  flight.
\newblock In {\em 2018 IEEE International Conference on Robotics and Automation
  (ICRA)}, pages 1--7. IEEE.

\bibitem[Mur-Artal et~al., 2015]{mur2015orb}
Mur-Artal, R., Montiel, J. M.~M., and Tardos, J.~D. (2015).
\newblock Orb-slam: a versatile and accurate monocular slam system.
\newblock {\em IEEE transactions on robotics}, 31(5):1147--1163.

\bibitem[Murali et~al., 2019]{Murali2019}
Murali, V., Spasojevic, I., Guerra, W., and Karaman, S. (2019).
\newblock {Perception-aware trajectory generation for aggressive quadrotor
  flight using differential flatness}.
\newblock In {\em 2019 American Control Conference (ACC)}, pages 3936--3943.
  IEEE.

\bibitem[Rojas-Perez and Martinez-Carranza, 2020]{rojas2020deeppilot}
Rojas-Perez, L.~O. and Martinez-Carranza, J. (2020).
\newblock Deeppilot: A cnn for autonomous drone racing.
\newblock {\em Sensors}, 20(16):4524.

\bibitem[Ronneberger et~al., 2015]{Ronneberger2015a}
Ronneberger, O., Fischer, P., and Brox, T. (2015).
\newblock {U-net: Convolutional networks for biomedical image segmentation}.
\newblock In {\em Lecture Notes in Computer Science (including subseries
  Lecture Notes in Artificial Intelligence and Lecture Notes in
  Bioinformatics)}, volume 9351, pages 234--241. Springer.

\bibitem[Sanket et~al., 2020]{Sanket2020}
Sanket, N.~J., Singh, C.~D., Ferm{\"{u}}ller, C., and Aloimonos, Y. (2020).
\newblock {PRGFlow: Benchmarking SWAP-Aware Unified Deep Visual Inertial
  Odometry}.
\newblock {\em arxiv}.

\bibitem[Scheper and de~Croon, 2016]{scheper2016abstraction}
Scheper, K.~Y. and de~Croon, G.~C. (2016).
\newblock Abstraction as a mechanism to cross the reality gap in evolutionary
  robotics.
\newblock In {\em International Conference on Simulation of Adaptive Behavior},
  pages 280--292. Springer.

\bibitem[Schmidhuber, 2015]{schmidhuber2015deep}
Schmidhuber, J. (2015).
\newblock Deep learning in neural networks: An overview.
\newblock {\em Neural networks}, 61:85--117.

\bibitem[Silver et~al., 2016]{silver2016mastering}
Silver, D., Huang, A., Maddison, C.~J., Guez, A., Sifre, L., Van Den~Driessche,
  G., Schrittwieser, J., Antonoglou, I., Panneershelvam, V., Lanctot, M.,
  et~al. (2016).
\newblock Mastering the game of go with deep neural networks and tree search.
\newblock {\em nature}, 529(7587):484--489.

\bibitem[Sutton and Barto, 2018]{sutton2018reinforcement}
Sutton, R.~S. and Barto, A.~G. (2018).
\newblock {\em Reinforcement learning: An introduction}.
\newblock MIT press.

\bibitem[Thrun et~al., 2006]{thrun2006stanley}
Thrun, S., Montemerlo, M., Dahlkamp, H., Stavens, D., Aron, A., Diebel, J.,
  Fong, P., Gale, J., Halpenny, M., Hoffmann, G., et~al. (2006).
\newblock Stanley: The robot that won the darpa grand challenge.
\newblock {\em Journal of field Robotics}, 23(9):661--692.

\bibitem[Vinyals et~al., 2019]{vinyals2019grandmaster}
Vinyals, O., Babuschkin, I., Czarnecki, W.~M., Mathieu, M., Dudzik, A., Chung,
  J., Choi, D.~H., Powell, R., Ewalds, T., Georgiev, P., et~al. (2019).
\newblock Grandmaster level in starcraft ii using multi-agent reinforcement
  learning.
\newblock {\em Nature}, 575(7782):350--354.

\bibitem[Yang et~al., 2018]{yang2018grand}
Yang, G.-Z., Bellingham, J., Dupont, P.~E., Fischer, P., Floridi, L., Full, R.,
  Jacobstein, N., Kumar, V., McNutt, M., Merrifield, R., et~al. (2018).
\newblock The grand challenges of science robotics.
\newblock {\em Science robotics}, 3(14):eaar7650.

\end{thebibliography}

\section*{Supplementary Material}

\subsection*{Data S1.}
Logfile of the winning championship heat \cite{AIRR_LOGS_2019}.

\subsection*{Movie S1}
Summary of the approach: {https://youtu.be/yN5QVl07F2Q}


\end{document}